\documentclass[manuscript,screen]{acmart}

\usepackage{multirow}
\usepackage{booktabs}
\usepackage{pifont}
\usepackage{longtable}
\usepackage{pdflscape}
\usepackage{afterpage}
\usepackage{caption}
\usepackage{subcaption}
\usepackage{forest}
\usepackage{adjustbox}
\usepackage{xcolor}
\usepackage{tabularx}
\usepackage{amsmath}
\usepackage{array}

\newcommand{\mycomment}[1]{\textcolor{red}{\textbf{[Mahsa: #1]}}}
\renewcommand{\mycomment}[1]{#1}

\AtBeginDocument{%
  }

\setcopyright{acmcopyright}
\copyrightyear{2023}
\acmYear{2023}
\acmDOI{XXXXXXX.XXXXXXX}

\begin{document}

\title{Deep Learning for Time Series Anomaly Detection: A Survey}

\author{Zahra Zamanzadeh Darban}
\email{zahra.zamanzadeh@monash.edu}
\orcid{0000-0003-2073-8072}
\authornotemark[1]
\affiliation{%
  \institution{Monash University}
  \city{Melbourne}
  \state{Victoria}
  \country{Australia}
}
\author{Geoffrey I.~Webb}
\email{geoff.webb@monash.edu}
\orcid{0000-0001-9963-5169}
\affiliation{%
  \institution{Monash University}
  \city{Melbourne}
  \state{Victoria}
  \country{Australia}
}
\author{Shirui Pan}
\email{s.pan@griffith.edu.au}
\orcid{0000-0003-0794-527X}
\affiliation{%
  \institution{Griffith University}
  \city{Gold Coast}
  \state{Queensland}
  \country{Australia}
}
\author{Charu C. Aggarwal}
\email{charu@us.ibm.com}
\orcid{0000-0003-2579-7581}
\affiliation{%
  \institution{IBM T. J. Watson Research Center}
  \city{Yorktown Heights}
  \state{NY}
  \country{USA}
}
\author{Mahsa Salehi}
\email{mahsa.salehi@monash.edu}
\orcid{0000-0002-2991-1612}
\affiliation{%
  \institution{Monash University}
  \city{Melbourne}
  \state{Victoria}
  \country{Australia}
}

\renewcommand{\shortauthors}{Darban et al.}

\begin{abstract}
Time series anomaly detection is important for a wide range of research fields and applications, including financial markets, economics, earth sciences, manufacturing, and healthcare. The presence of anomalies can indicate novel or unexpected events, such as production faults, system defects, and heart palpitations, and is therefore of particular interest. The large size and complexity of patterns in time series data have led researchers to develop specialised deep learning models for detecting anomalous patterns. This survey provides a structured and comprehensive overview of state-of-the-art deep learning for time series anomaly detection. It provides a taxonomy based on anomaly detection strategies and deep learning models. Aside from describing the basic anomaly detection techniques in each category, their advantages and limitations are also discussed. Furthermore, this study includes examples of deep anomaly detection in time series across various application domains in recent years. Finally, it summarises open issues in research and challenges faced while adopting deep anomaly detection models to time series data.
\end{abstract}



\maketitle

\section{Introduction}
The detection of anomalies, also known as outlier or novelty detection, has been an active research field in numerous application domains since the 1960s \citep{grubbs1969procedures}. As computational processes evolve, the collection of big data and its use in artificial intelligence (AI) is better enabled, contributing to time series analysis including the detection of anomalies. With greater data availability and increasing algorithmic efficiency/computational power, time series analysis is increasingly used to address business applications through forecasting, classification, and anomaly detection \citep{esling2012time}, \citep{carreno2020analyzing}. Time series anomaly detection (TSAD) has received increasing attention in recent years, because of increasing applicability in a wide variety of domains, including urban management, intrusion detection, medical risk, and natural disasters.

Deep learning has become increasingly capable over the past few years of learning expressive representations of complex time series, like multidimensional data with both spatial (intermetric) and temporal characteristics. In deep anomaly detection, neural networks are used to learn feature representations or anomaly scores in order to detect anomalies. Many deep anomaly detection models have been developed, providing significantly higher performance than traditional time series anomaly detection tasks in different real-world applications. 

Although the field of anomaly detection has been explored in several literature surveys \citep{chandola2009anomaly}, \citep{pang2021deep}, \citep{chalapathy2019deep}, \citep{blazquez2021review}, \citep{braei2020anomaly} and some evaluation review papers exist \citep{schmidl2022anomaly}, \citep{kim2022towards}, there is only one survey on deep anomaly detection methods for time series data \citep{choi2021deep}. However, the mentioned survey \citep{choi2021deep} has not covered the vast range of TSAD methods that have emerged in recent years, such as DAEMON \citep{chen2021daemon}, TranAD \citep{tuli2022tranad}, DCT-GAN \citep{li2021dct}, and Interfusion \citep{li2021multivariate}. Additionally, the representation learning methods within the taxonomy of TSAD methodologies has not been addressed in this survey. As a result, there is a need for a survey that enables researchers to identify important future directions of research in TSAD and the methods that are suitable to various application settings. Specifically, this article makes the following contributions:
\begin{itemize}
    \item \textbf{Taxonomy}: We present a novel taxonomy of deep anomaly detection models for time series data. These models are broadly classified into four categories: forecasting-based, reconstruction-based, representation-based and hybrid methods. Each category is further divided into subcategories based on the deep neural network architectures used. This taxonomy helps to characterise the models by their unique structural features and their contribution to anomaly detection capabilities.
    
    \item \textbf{Comprehensive Review}: Our study provides a thorough review of the current state-of-the-art in time series anomaly detection up to 2024. This review offers a clear picture of the prevailing directions and emerging trends in the field, making it easier for readers to understand the landscape and advancements.
    
    \item \textbf{Benchmarks and Datasets}: We compile and describe the primary benchmarks and datasets used in this field. Additionally, we categorise the datasets into a set of domains and provide hyperlinks to these datasets, facilitating easy access for researchers and practitioners.

    \item \textbf{Guidelines for Practitioners}: Our survey includes practical guidelines for readers on selecting appropriate deep learning architectures, datasets, and models. These guidelines are designed to assist researchers and practitioners in making informed choices based on their specific needs and the context of their work.
    
    \item \textbf{Fundamental Principles}: We discuss the fundamental principles underlying the occurrence of different types of anomalies in time series data. This discussion aids in understanding the nature of anomalies and how they can be effectively detected.
    
    \item \textbf{Evaluation Metrics and Interpretability}: We provide an extensive discussion on evaluation metrics together with guidelines for metric selection. Additionally, we include a detailed discussion on model interpretability to help practitioners understand and explain the behaviour and decisions of TSAD models.
\end{itemize}

This article is organised as follows. In Section \ref{sec:TS}, we start by introducing preliminary definitions, which is followed by a taxonomy of anomalies in time series. Section \ref{sec:DAD} discusses the application of deep anomaly detection models to time series. Different deep models and their capabilities are then presented based on the main approaches (forecasting-based, reconstruction-based, representation-based, and hybrid) and architectures of deep neural networks. Additionally, Section \ref{sec:APP} explores the applications of time series deep anomaly detection models in different domains. Finally, Section \ref{sec:Discuss} provides several challenges in this field that can serve as future opportunities. An overview of publicly available and commonly used datasets for the considered anomaly detection models can be found in Section \ref{sec:Dataset}.

\section{Background} \label{sec:TS}
A time series is a series of data points indexed sequentially over time. 
The most common form of time series is a sequence of observations recorded over time \citep{hamilton2020time}.
Time series are often divided into \emph{univariate} (one-dimensional) and \emph{multivariate} (multi-dimensional).
These two types are defined in the following subsections.
Thereafter, decomposable components of the time series are outlined. Following that, we provide a taxonomy of anomaly types based on time series' components and characteristics.

\subsection{Univariate Time Series}
As the name implies, a univariate time series (UTS) is a series of data that is based on a single variable that changes over time, as shown in Fig. \ref{fig:temporal}. Keeping a record of the humidity level every hour of the day would be an example of this. The time series $X$ with $t$ timestamps can be represented as an ordered sequence of data points in the following way:
\begin{equation}
    X = (x_1, \ x_2,\ldots, \ x_t)
\end{equation}
where $x_i$ represents the data at timestamp $i \in T$ and $T = \{1, 2, ..., t\}$. 

\subsection{Multivariate Time Series}
Additionally, a multivariate time series (MTS) represents multiple variables that are dependent on time, each of which is influenced by both past values (stated as ``temporal'' dependency) and other variables (dimensions) based on their correlation. The correlations between different variables are referred to as spatial or intermetric dependencies in the literature, and they are used interchangeably \citep{li2021multivariate}. In the same example, air pressure and temperature would also be recorded every hour besides humidity level.

An example of an MTS with two dimensions is illustrated in Fig. \ref{fig:intermetric}. Consider a MTS represented as a sequence of vectors over time, each vector at time $i$, $X_i$, consisting of $d$ dimensions:
\begin{equation}
X = (X_1, X_2, \ldots, X_t) = ((x^1_1, x^2_1, \ldots, x^d_1), (x^1_2, x^2_2, \ldots, x^d_2), \ldots, (x^1_t, x^2_t, \ldots, x^d_t))
\end{equation}
where $X_i = (x^1_i, x^2_i, \ldots, x^d_i)$ represents a data vector at time $i$, with each $x^j_i$ indicating the observation at time $i$ for the $j^{th}$ dimension, and $j = {1, 2, \ldots, d}$, where $d$ is the total number of dimensions.

\subsection{Time Series Decomposition}
It is possible to decompose a time series $X$ into four components, each of which express a specific aspect of its movement \citep{ref1}. The components are as follows:
\begin{itemize}
\item {\texttt{Secular trend}}: This is the long-term trend in the series, such as increasing, decreasing or stable. The secular trend represents the general pattern of the data over time and does not have to be linear. The change in population in a particular region over several years is an example of nonlinear growth or decay depending on various dynamic factors.
\item {\texttt{Seasonal variations}}: Depending on the month, weekday, or duration, a time series may exhibit a seasonal pattern. Seasonality always occurs at a fixed frequency. For instance, a study of gas/electricity consumption shows that the consumption curve does not follow a similar pattern throughout the year. Depending on the season and the locality, the pattern is different.
\item {\texttt{Cyclical fluctuations}}: A cycle is defined as an extended deviation from the underlying series defined by the secular trend and seasonal variations. Unlike seasonal effects, cyclical effects vary in onset and duration. Examples include economic cycles such as booms and recessions.
\item {\texttt{Irregular variations}}: This refers to random, irregular events. It is the residual after all the other components are removed. A disaster such as an earthquake or flood can lead to irregular variations.
\end{itemize}
A time series can be mathematically described by estimating its four components separately, and each of them may deviate from the normal behaviour. 

\subsection{Anomalies in Time Series} \label{sub:anoms}
According to \citep{hawkins1980identification}, the term anomaly refers to a deviation from the general distribution of data, such as a single observation (point) or a series of observations (subsequence) that deviate greatly from the general distribution. A small portion of the dataset contains anomalies, indicating the dataset mostly follows a normal pattern. There may be considerable amounts of noise embedded in real-world data, and such noise may be irrelevant to the researcher \citep{aggarwal2017introduction}. The most meaningful deviations are usually those that are significantly different from the norm. In circumstances where noise is present, the main characteristics of the data are identical.
In data domains such as time series, trend analysis and anomaly detection are closely related, but they are not equivalent \citep{aggarwal2017introduction}. It is possible to see changes in time series datasets owing to concept drift, which occurs when values and trends change over time gradually or abruptly \citep{masud2010addressing}, \citep{aggarwal2007data}. 
\subsubsection{Types of Anomalies}
\begin{figure}[t]
  \centering
  \begin{subfigure}[b]{0.49\textwidth}
    \centering
    \includegraphics[width=\textwidth]{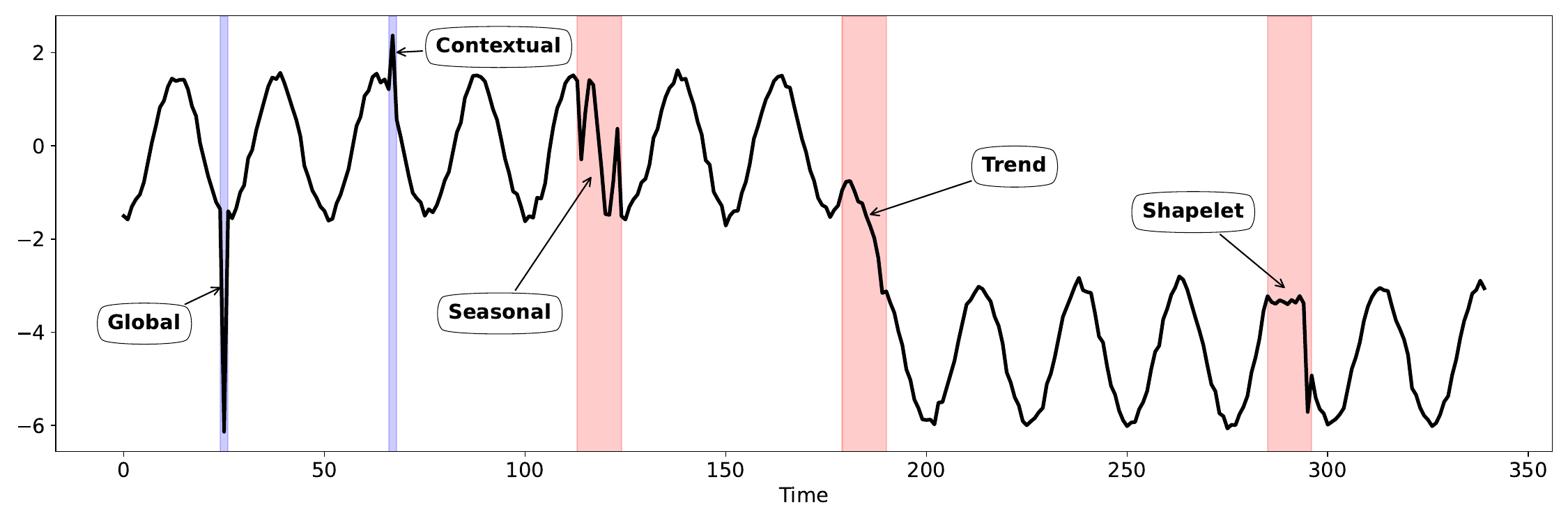}
    \caption{}
    \label{fig:temporal}
  \end{subfigure}
  \hfill
  \begin{subfigure}[b]{0.49\textwidth}
    \centering
    \includegraphics[width=\textwidth]{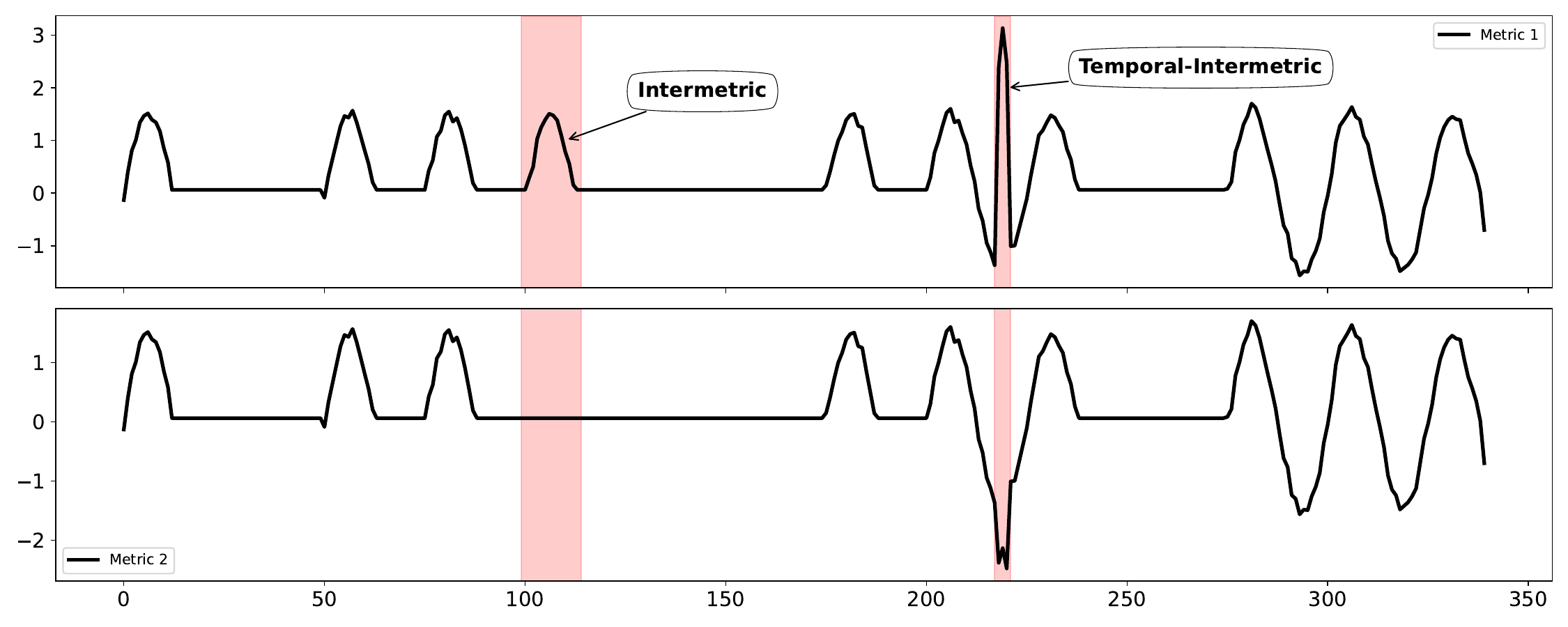}
    \caption{}
    \label{fig:intermetric}
  \end{subfigure}
  \caption{(a) An overview of different temporal anomalies plotted from the NeurIPS-TS dataset \citep{lai2021revisiting}. Global and contextual anomalies occur in a point (coloured in blue). Seasonal, trend and shapelet can occur in a subsequence (coloured in red). (b) Intermetric and temporal-intermetric anomalies in MTS. In this figure, metric 1 is power consumption, and metric 2 is CPU usage.}
  \vspace{-10pt}
\end{figure}


Anomalies in UTS and MTS can be classified as temporal, intermetric, or temporal-intermetric anomalies \citep{li2021multivariate}. In a time series, temporal anomalies can be compared with either their neighbours (local) or the whole time series (global), and they present different forms depending on their behaviour \citep{lai2021revisiting}. There are several types of temporal anomalies that commonly occur in UTS, all of which are shown in Fig. \ref{fig:temporal}. Temporal anomalies can also occur in the MTS and affect multiple dimensions or all dimensions. A subsequent anomaly may appear when an unusual pattern of behaviour emerges over time; however, each observation may not be considered an outlier by itself. As a result of a point anomaly, an unexpected event occurs at one point in time, and it is assumed to be a short sequence.
Different types of temporal anomalies are as follows:
\begin{itemize}

\item {\texttt{Global}}: They are spikes in the series, which are point(s) with extreme values compared to the rest of the series. A global anomaly, for instance, is an unusually large payment by a customer on a typical day. Considering a threshold, it can be described as Eq. (\ref{eq:global}).
\begin{equation}
\label{eq:global}
    | x_t - \hat{x}_t| > threshold
\end{equation}
where $\hat{x}$ is the output of the model. If the difference between the output and actual point value is greater than a threshold, then it has been recognised as an anomaly. An example of a global anomaly is shown on the left side of Fig. \ref{fig:temporal} where $-6$ has a large deviation from the time series.

\item {\texttt{Contextual}}: A deviation from a given context is defined as a deviation from a neighbouring time point, defined here as one that lies within a certain range of proximity. These types of anomalies are small glitches in sequential data, which are deviated values from their neighbours. It is possible for a point to be normal in one context while an anomaly in another. For example, large interactions, such as those on boxing day, are considered normal, but not so on other days.
The formula is the same as that of a global anomaly, but the threshold for finding anomalies differs. The threshold is determined by taking into account the context of neighbours:
\begin{equation}
    threshold\ \approx \lambda \ \times \ \mathbf{var}(X_{t-w: t})
\end{equation}
where $X_{t-w: t}$ refers to the context of the data point $x_t$ with a window size $w$, $\mathbf{var}$ is the variance of the context of data point and $\lambda$ controlling coefficient for the threshold. The second blue highlight in Fig. \ref{fig:temporal} is a contextual anomaly that occurs locally in a specific context.

\item {\texttt{Seasonal}}: In spite of normal shapes and trends of the time series, their seasonality is unusual compared to the overall seasonality. An example is the number of customers in a restaurant during a week. Such a series has a clear weekly seasonality, so it makes sense to look for deviations in this seasonality and process the anomalous periods individually.
\begin{equation}
    diss_s(S ,\ \hat{S}) > threshold
\end{equation}
where $diss_s$ is a function measuring the dissimilarity between two subsequences and $\hat{S}$ denotes the seasonality of the expected subsequences. As demonstrated in the first red highlight of Fig. \ref{fig:temporal}, the seasonal anomaly changes the frequency of a rise and drop of data in the particular segment.

\item {\texttt{Trend}}: The event that causes a permanent shift in the data to its mean and produces a transition in the trend of the time series. While this anomaly preserves its cycle and seasonality of normality, it drastically alters its slope. Trends can occasionally change direction, meaning they may go from increasing to decreasing and vice versa. As an example, when a new song comes out, it becomes popular for a while, then it disappears from the charts like the segment in Fig. \ref{fig:temporal} where the trend is changed and is assumed as a trend anomaly. It is likely that the trend will restart in the future.
\begin{equation}
    diss_t(T ,\ \hat{T}) > threshold
\end{equation}
where $\hat{T}$ is the normal trend. 

\item {\texttt{Shapelet}}: Shapelet means a distinctive, time series subsequence pattern. There is a subsequence whose time series pattern or cycle differs from the usual pattern found in the rest of the sequence. Variations in economic conditions, like the total demand for and supply of goods and services, are often the cause of these fluctuations. In the short-run, these changes lead to periods of expansion and recession.
\begin{equation}
    diss_c(C ,\ \hat{C}) > threshold
\end{equation}
where $\hat{C}$ specifies the cycle or shape of expected subsequences. For example, the last highlight in Fig.~\ref{fig:temporal} where the shape of the segment changed due to some fluctuations.

\end{itemize}
Having discussed various types of anomalies, we understand that these can often be characterised by the distance between the actual subsequence observed and the expected subsequence. In this context, dynamic time warping (DTW) \citep{muller2007dynamic}, which optimally aligns two time series, is a valuable method for measuring this dissimilarity. Consequently, DTW's ability to accurately calculate temporal alignments makes it a suitable tool for anomaly detection applications, as evidenced in several studies \citep{benkabou2018unsupervised}, \citep{song2022robust}. 
Moreover, MTS is composed of multiple dimensions (a.k.a, metrics \citep{su2019robust, li2021multivariate}) that each describe a different aspect of a complex entity. Spatial dependencies (correlations) among dimensions within an entity, also known as intermetric dependencies, can be linear or nonlinear. MTS would exhibit a wide range of anomalous behaviour if these correlations were broken. An example is shown in the left part of Fig. \ref{fig:intermetric}. The correlation between power consumption in the first dimension (metric 1) and CPU usage in the second dimension (metric 2) usage is positive, but it breaks about 100th of a second after it begins. Such an anomaly is named the intermetric anomaly in this study. 

\begin{equation}
\max_{\forall j, k \in D, j \neq k} \text{diss}_{\text{corr}}(\mathbf{Corr}(X^{j}, X^{k}), \mathbf{Corr}(X^{j}_{t+\delta t_j:t+w+\delta t_j}, X^{k}_{t+\delta t_k:t+w+\delta t_k})) > \text{threshold}
\end{equation}
where
$X^{j}$ and $X^{k}$ are different dimensions of the MTS,
$\mathbf{Corr}$ denotes the correlation function that measures the relationship between two dimensions,
$\delta t_j$ and $\delta t_k$ are time shifts that adjust the comparison windows for dimensions $j$ and $k$, accommodating asynchronous events or delays between observations,
$t$ is the starting point of the time window,
$w$ is the width of the time window, indicating the duration over which correlations are assessed,
$\text{diss}_{\text{corr}}$ is a function that quantifies the divergence in correlation between the standard, long-term measurement and the dynamic, short-term measurement within the specified window,
$\text{threshold}$ is a predefined limit that determines when the divergence in correlations signifies an anomaly,
and $D$ is the set of all dimensions within the MTS, with the comparison conducted between every unique pair $(j, k)$ where $j \neq k$.

Dimensionality reduction techniques, such as selecting a subset of critical dimensions based on domain knowledge or preliminary analysis, help manage the computational complexity that increases with the number of dimensions.

Where $X^j$ and $X^k$ are two different dimensions of MTS that are correlated, and  $corr$ measures the correlations between two dimensions. When this correlation deteriorates in the window $t:t+w$, it means that the coefficient deviates more than  $threshold$ from the normal coefficient.

Intermetric-temporal anomalies introduce added complexity and challenges in anomaly detection; however, they occasionally facilitate easier detection across temporal or various dimensional perspectives due to their simultaneous violation of intermetric and temporal dependencies, as illustrated on the right side of Fig. \ref{fig:intermetric}.

\section{Time series Anomaly Detection Methods} \label{sec:DAD}

\mycomment{Traditional methods offer varied approaches to time series anomaly detection. \textit{Statistical-based} methods \citep{yamanishi2002unifying} aim to learn a statistical model of the normal behaviour of time series. In \textit{clustering-based} approaches \citep{moshtaghi2014evolving}, a normal profile of time series windows is learned, and the distance to the centroid of the normal clusters is considered as an anomaly score, or clusters with a small number of members are considered as anomaly clusters. \textit{Distances-based} approaches are extensively studied \citep{yeh2016matrix}, in which the distance of a window of time series to its nearest neighbours is considered as an anomaly score. \textit{Density-based} approaches \citep{ding2013anomaly} estimate the density of data points and time series windows with low density are detected as anomalies.}

\begin{figure}[t]
    \centering
    \begin{adjustbox}{max width=0.7\textwidth}
    \begin{forest}
    for tree={
        grow=south,
        parent anchor=south,
        child anchor=north,
        rectangle,
        draw,
        inner sep=5pt,
        align=center,
        baseline,
        l sep+=10pt,
        s sep+=10pt,
        edge path={
            \noexpand\path [draw, \forestoption{edge}]
            (!u.parent anchor) -- +(0,-7pt) -| (.child anchor)\forestoption{edge label};
        },
    }
    [Deep Learning Architectures, fill=gray!10
        [CNN, fill=blue!10
            [TCN, fill=green!10]
            [ResNet, fill=green!10]
        ]
        [RNN, fill=blue!10
            [LSTM, fill=green!10]
            [Bi-LSTM, fill=green!10]
            [GRU, fill=green!10]
        ]
        [AE, fill=blue!10
            [SAE, fill=green!10]
            [DAE, fill=green!10]
            [CAE, fill=green!10]
            [VAE, fill=green!10]
        ]
        [GNN, fill=blue!10
            [GCN, fill=green!10]
            [GAT, fill=green!10]
        ]
        [Transformer, fill=blue!10]
        [GAN, fill=blue!10]
        [HTM, fill=blue!10]
    ]
    \end{forest}
    \end{adjustbox}
    \caption{Deep Learning architectures used in time series anomaly detection}
    \label{fig:dl}
\end{figure}
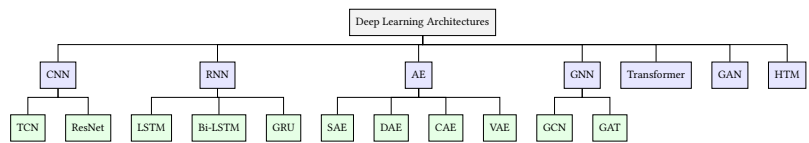

In data with complex structures, deep neural networks are powerful for modelling temporal and spatial dependencies in time series. A number of scholars have explored their application to anomaly detection using various deep architectures, as illustrated in Fig \ref{fig:dl}. 


\subsection{Deep Models for Time Series Anomaly Detection} \label{sub:TS-AD}
\begin{figure}[t]
  \centering
  \includegraphics[width=0.7\linewidth]{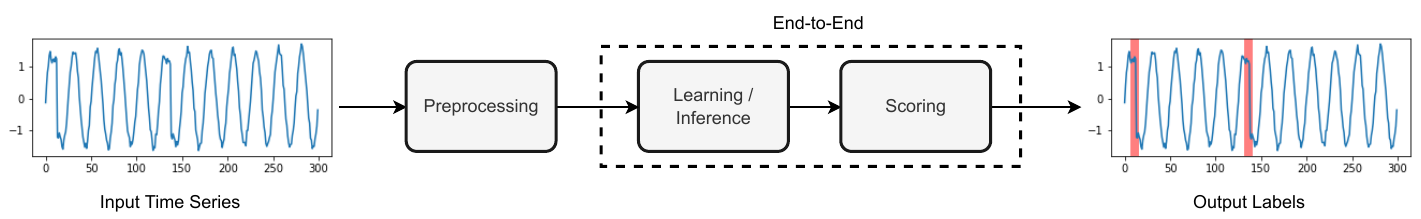}
  \caption{General components of deep anomaly detection models in time series}
  \label{fig:components}
  \vspace{-10pt}
\end{figure}

An overview of deep anomaly detection models in time series is shown in Fig. \ref{fig:components}. In our study, deep models for anomaly detection in time series are categorised based on their main approach and architectures. There are two main approaches (learning component in Fig. \ref{fig:components}) in the TSAD literature: \textit{forecasting-based} and \textit{reconstruction-based}. A forecasting-based model can be trained to predict the next time stamp, whereas a reconstruction-based model can be deployed to capture the embedding of time series data. A categorisation of deep learning architectures in TSAD is shown in Fig. \ref{fig:dl}.

The TSAD models are summarised in Table \ref{tab:uts-table} and Table \ref{tab:mts-table} based on the input dimensions they process, which are UTS and MT\mycomment{S}, respectively. These tables give an overview of the following aspects of the models: Temporal/Spatial, Learning scheme, Input, Interpretability, Point/Sub-sequence anomaly, Stochasticity, Incremental, and \mycomment{Univariate support}. However, Table \ref{tab:uts-table} excludes columns for Temporal/Spatial, Interpretability, and \mycomment{Univariate support} as these features pertain solely to MTS. Additionally, it lacks an Incremental column because no univariate models incorporate an incremental approach.

\subsubsection{Temporal/Spatial}
With a UTS as input, a model can capture temporal information (i.e., pattern), while with a MTS as input, it can learn normality through both temporal and spatial dependencies. Moreover, if the model input is an MTS in which spatial dependencies are captured, the model can also detect intermetric anomalies (shown in Fig. \ref{fig:intermetric}).

\subsubsection{Learning Schemes}
In practice, training data tends to have a very small number of anomalies that are labelled. As a consequence, most of the models attempt to learn the representation or features of normal data. Based on anomaly definitions, anomalies are then detected by finding deviations from normal data.
There are four learning schemes in the recent deep models for anomaly detection: unsupervised, supervised, semi-supervised, and self-supervised. These are based on the availability (or lack) of labelled data points. \textit{Supervised} method employs a distinct method of learning the boundaries between anomalous and normal data that is based on all the labels in the training set. It can determine an appropriate threshold value that will be used for classifying all timestamps as anomalous if the anomaly score (Section \ref{sub:TS-AD}) assigned to those timestamps exceeds the threshold.
The problem with this method is that it is not applicable to many real-world applications because anomalies are often unknown or improperly labelled.
In contrast, \textit{Unsupervised} approach uses no labels and makes no distinction between training and testing datasets. These techniques are the most flexible since they rely exclusively on intrinsic features of the data. They are useful in streaming applications because they do not require labels for training and testing. Despite these advantages, researchers may encounter difficulties evaluating anomaly detection models using unsupervised methods. The anomaly detection problem is typically treated as an unsupervised learning problem due to the inherently unlabelled nature of historical data and the unpredictable nature of anomalies.
\textit{Semi-supervised} anomaly detection in time series data may be utilised in cases where the dataset only consists of \mycomment{labelled normal data}, unlike supervised methods that require a fully labelled dataset of both normal and anomalous points. Unlike unsupervised methods, which detect anomalies without any labelled data, semi-supervised TSAD relies on labelled normal data to define normal patterns and detect deviations as anomalies. This approach is distinct from \textit{self-supervised} learning, where 
\mycomment{the model generates its own supervisory signal from the input data without needing explicit labels.}

\subsubsection{Input}
A model may take an individual point (i.e., a time step) or a window (i.e., a sequence of time steps containing historical information) as an input. Windows can be used in order, also called sliding windows, or shuffled without regard to the order, depending on the application. To overcome the challenges of comparing subsequences rather than points, many models use representations of subsequences (windows) instead of raw data and employ sliding windows that contain the history of previous time steps that rely on the order of subsequences within the time series data. A sliding window extraction is performed in the preprocessing phase after other operations have been implemented, such as imputing missing values, downsampling or upsampling of the data, and data normalisation.

\subsubsection{Interpretability}
In interpretation, the cause of an anomalous observation is given. Interpretability is essential when anomaly detection is used as a diagnostic tool since it facilitates troubleshooting and analysing anomalies. MTS are challenging to interpret, and stochastic deep learning complicates the process even further. A typical procedure to troubleshoot entity anomalies involves searching for the top dimension that differs most from previously observed behaviour. In light of that, it is, therefore, possible to interpret a detected entity anomaly by analysing several dimensions with the highest anomaly scores.

\subsubsection{Point/Subsequence anomaly}
The model can detect either point anomalies or subsequence anomalies. A point anomaly is a point that is unusual when compared with the rest of the dataset. Subsequence anomalies occur when consecutive observations have unusual cooperative behaviour, although each observation is not necessarily an outlier on its own. Different types of anomalies are described in Section \ref{sub:anoms} and illustrated in Fig. \ref{fig:temporal} and Fig. \ref{fig:intermetric}

\subsubsection{Stochasticity}
As shown in Tables \ref{tab:uts-table} and  \ref{tab:mts-table}, we investigate the stochasticity of anomaly detection models as well. Deterministic models can accurately predict future events without relying on randomness. Predicting something that is deterministic is easy because you have all the necessary data at hand. The models will produce the same exact results for a given set of inputs in this circumstance. Stochastic models can handle uncertainties in the inputs. Through the use of a random component as an input, you can account for certain levels of unpredictability or randomness.

\subsubsection{Incremental}
This is a machine learning paradigm in which the model's knowledge extends whenever one or more new observations appear. It specifies a dynamic learning strategy that can be used if training data becomes available gradually. The goal of incremental learning is to adapt a model to new data while preserving its past knowledge.

\begin{table}[h]
\caption{Univariate Deep Anomaly Detection Models in Time Series}
\label{tab:uts-table}
\centering
\small
\resizebox{0.6\columnwidth}{!}{
\renewcommand{\arraystretch}{1.2}
\begin{tabular}{@{}l|l|llllll@{}}
\toprule
A$^1$                           & MA$^2$               & Model  & Year  & Su/Un$^3$ & Input$^4$ & P/S$^5$ & Stc$^6$        \\ \midrule
\multirow{7}{*}{\rotatebox[origin=c]{90}{Forecasting}}   
    & \multirow{5}{*}{RNN (\ref{sub:for-rnn})}  & LSTM-AD  \citep{malhotra2015long}          & 2015 & Un        & P     & Point          \\
    &     & DeepLSTM \citep{chauhan2015anomaly} & 2015      & Semi      & P     &  Point     &         \\
    &                      & LSTM RNN  \citep{bontemps2016collective}   & 2016 & Semi      & P     & Subseq         \\
    &                      & LSTM-based  \citep{ergen2019unsupervised}  & 2019 & Un        & W     & -              \\
    &                      & TCQSA  \citep{liu2020anomaly}   &           2020 & Su       & P    & -              \\ \cmidrule(l){2-8}
    & \multirow{2}{*}{HTM (\ref{sub:for-htm})}  & Numenta HTM  \citep{ahmad2017unsupervised} & 2017 & Un       & -     & -              \\
    &                      & Multi HTM  \citep{wu2018hierarchical}      & 2018 & Un        & -     & -              \\ \cmidrule(l){2-8}
    & CNN  (\ref{sub:for-cnn})                & SR-CNN  \citep{ren2019time}              & 2019 & Un        & W     & Point + Subseq \\ \midrule
\multirow{4}{*}{\rotatebox[origin=c]{90}{Reconstruction}} 
    & \multirow{3}{*}{VAE (\ref{sub:rec-vae})}  & Donut  \citep{xu2018unsupervised}          & 2018 & Un        & W     & Subseq & \ding{51}
    \\
    &                      & Bagel  \citep{li2018robust}                & 2018 & Un        & W     & Subseq & \ding{51} \\
    &                      & Buzz  \citep{chen2019unsupervised}         & 2019 & Un        & W     & Subseq & \ding{51}                 \\ \cmidrule(l){2-8}
    & AE  (\ref{sub:rec-ae})               & EncDec-AD  \citep{malhotra2016lstm}        & 2016 & Semi      & W     & Point          \\
    \bottomrule
\end{tabular}
}
\\
\tiny\raggedright $^1$ A: \textit{Approach}, 
$^2$ MA: \textit{Main Architecture},
$^3$ Su/Un: \textit{Supervised/Unsupervised} | Values: [Su: \textit{Supervised}, Un: \textit{Unsupervised}, Semi: \textit{Semi-supervised}, Self: \textit{Self-supervised}],
$^4$ Input: \textit{P: point / W: window},
$^5$ P/S: \textit{Point/Sub-sequence}, $^6$ Stc: \textit{Stochastic}, $"-"$ \textit{indicates a feature is not defined or mentioned.}
\vspace{-10pt}
\end{table}

\begin{table}[]
\caption{Multivariate Deep Anomaly Detection Models in Time Series}
\label{tab:mts-table}
\resizebox{0.85\columnwidth}{!}{
\begin{tabular}{@{}l|l|llllllllll@{}}
\toprule
A$^1$ & MA$^2$ & Model & Year & T/S$^3$ & Su/Un$^4$ & Input$^5$ & Int$^6$ & P/S$^7$ & Stc$^8$ & Inc$^9$ & US$^{10}$                    
\\ \midrule
\multirow{14}{*}{\rotatebox[origin=c]{90}{Forecasting}}    
& \multirow{5}{*}{RNN (\ref{sub:for-rnn})}   
      & LSTM-PRED \citep{goh2017anomaly} & 2017         & T       & Un        & W     & \ding{51} & -              &                            &                            \\
    & & LSTM-NDT \citep{hundman2018detecting} & 2018  & T       & Un        & W     & \ding{51} & Subseq         &                            &                \\
     & & LGMAD \citep{ding2019real} & 2019    & T       & Semi      & P     &         & Point    &    &   & \ding{51}                          \\
     &  & THOC \citep{shen2020timeseries} & 2020         & T       & Self      & W     &                            & Subseq         &                    & & \ding{51}                            \\
     &  & AD-LTI \citep{wu2020developing} & 2020      & T       & Un        & P     &                            & Point (frame)  &        &       \\ \cmidrule(l){2-12}
    & \multirow{3}{*}{CNN (\ref{sub:for-cnn})}              
        & DeepAnt \citep{munir2018deepant} & 2018 & T       & Un      & W     &                            & Point + Subseq         &        & \\
        & & TCN-ms \citep{he2019temporal}  & 2019      & T       & Semi      & W     &     & Subseq   &    & & \ding{51} \\
        & & TimesNet \citep{wu2023timesnet}  & 2023  & T & Un  & W &     & - &    &  & \ding{51}\\
                \cmidrule(l){2-12} 
     & \multirow{3}{*}{GNN (\ref{sub:for-gnn})}         
     & GDN \citep{deng2021graph} & 2021 & S       & Un        & W     & \ding{51} & -              &                            &                            \\
     &  & GTA* \citep{chen2021learning} & 2021        & ST      & Semi      & -     &                            & -              &                            &                            \\
     &  & GANF \citep{dai2022graph} & 2022      & ST      & Un        & W     &       &      &        &                            \\ \cmidrule(l){2-12} 
     & HTM (\ref{sub:for-htm})                         
     & RADM \citep{ding2018multivariate} &  2018     & T       & Un        & W     &                            & -              &                            &                            \\ \cmidrule(l){2-12} 
     & \multirow{2}{*}{Transformer (\ref{sub:for-tran})} 
     & SAND \citep{song2018attend} & 2018 & T       & Semi       & W  &    & -   &      & \\
     &  & GTA* \citep{chen2021learning} & 2021 & ST      & Semi      & -     &    & -    &     &   
     
     \\ \midrule
     
\multirow{30}{*}{\rotatebox[origin=c]{90}{Reconstruction}} & \multirow{7}{*}{AE (\ref{sub:rec-ae})}          
        & AE/DAE \citep{sakurada2014anomaly} & 2014     & T       & Semi      & P     &   & Point   &    &     \\
         &  & DAGMM \citep{zong2018deep} & 2018 & S       & Un        & P     &                            & Point          & \ding{51} &              \\
         &  & MSCRED \citep{zhang2019deep} & 2019          & ST      & Un        & W     & \ding{51} & Subseq         &                            &                            \\
         & & USAD \citep{audibert2020usad} & 2020         & T       & Un        & W     &    & Point   &   &          \\
         &  & APAE \citep{goodge2020robustness} & 2020      & T       & Un        & W     &        & -      &   &     \\
         &  & RANSynCoders \citep{abdulaal2021practical} & 2021 & ST      & Un        & P     & \ding{51} & Point          &                            & \ding{51} \\
         & & CAE-Ensemble \citep{campos2021unsupervised}& 2021 & T       & Un        & W     &     & Subseq         &                            &           \\
         &  & AMSL \citep{zhang2022adaptive} & 2022  & T       & Self      & W     &       & -        &        &                           
         \\ 
          &  & ContextDA \citep{lai2023context} & 2023  & T & Un   & W     &       & Point + Subseq  &        &       
         \\
         \cmidrule(l){2-12} 
          & \multirow{12}{*}{VAE (\ref{sub:rec-vae})}  
          & STORN \citep{solch2016variational} & 2016    & ST      & Un        & P     &                            & Point          & \ding{51} &      \\
         &  & GGM-VAE \citep{guo2018multidimensional} & 2018  & T       & Un        & W     &         & Subseq         &  \ding{51}  & 
         \\
         & & LSTM-VAE \citep{park2018multimodal} & 2018    & T       & Semi        & P     &              & -              &  \ding{51}  &                \\
         &  & OmniAnomaly \citep{su2019robust} &   2019     & T       & Un        & W     & \ding{51} & Point + Subseq & \ding{51} &    \\
          &   & VELC \citep{zhang2019velc} & 2019            & T       & Un        & -     &                            & -    &  \ding{51}   &            \\
         &  & SISVAE \citep{li2020anomaly} & 2020 & T  & Un        & W  &  & Point   & \ding{51}  &   & \ding{51} \\
         &   & VAE-GAN \citep{niu2020lstm} & 2020     & T       & Semi      & W     &                            & Point     & \ding{51}  &   &  \ding{51}\\
         &  & TopoMAD \citep{he2020spatiotemporal} & 2020  & ST      & Un        & W     &            & Subseq         & \ding{51} &                            \\
         &  & PAD \citep{chen2021joint} & 2021   & T       & Un        & W     &                            & Subseq         &  \ding{51} &      &  \ding{51}\\
         &   & InterFusion \citep{li2021multivariate} & 2021 & ST      & Un        & W     & \ding{51} & Subseq         &  \ding{51}  &                            \\
         &  & MT-RVAE* \citep{wang2022variational} & 2022 & ST      & Un        & W     &                            & -              &  \ding{51} &                            \\
         & & RDSMM \citep{li2022learning} & 2022  & T       & Un     & W &  & Point + Subseq & \ding{51} & & \ding{51}
         \\ \cmidrule(l){2-12}   
         & \multirow{5}{*}{GAN (\ref{sub:rec-gan})}         
         & MAD-GAN \citep{li2019mad} & 2019           & ST      & Un        & W     &              & Subseq         &                            &               \\
         &  & BeatGAN \citep{zhou2019beatgan} & 2019    & T       & Un  & W &    & Subseq &   &  & \ding{51}             \\
         &  & DAEMON \citep{chen2021daemon} & 2021       & T       & Un        & W     & \ding{51} & Subseq         &                            &                     \\
         &  & FGANomaly \citep{du2021gan} & 2021  & T       & Un        & W     &                            & Point + Subseq &      &                            \\
         &  & DCT-GAN* \citep{li2021dct} & 2021   & T       & Un        & W     &   & -  &  &    & \ding{51} 
         \\ \cmidrule(l){2-12} 
         & \multirow{5}{*}{Transformer (\ref{sub:rec-tran})} 
         & Anomaly Transformer \citep{xu2021anomaly} & 2021 & T       & Un        & W     &           & Subseq         &                            &                \\
          & & DCT-GAN* \citep{li2021dct} & 2021 & T       & Un        & W     &       & -   &  &    & \ding{51}\\
         &  & TranAD \citep{tuli2022tranad} & 2022     & T       & Un        & W     & \ding{51} & Subseq         &                            & & \ding{51}                           \\
         & & MT-RVAE* \citep{wang2022variational} & 2022    & ST      & Un        & W     &           & -              &          &             \\ 
          & & Dual-TF \citep{nam2024breaking} & 2024 & T & Un & W &           & Point + Subseq &  &  & \ding{51} \\ 
         \midrule
         
    \multirow{5}{*}{\rotatebox[origin=c]{90}{Representation}}          & \multirow{1}{*}{Transformer (\ref{sub:rep-tran})}          
        & TS2Vec \citep{yue2022ts2vec} & 2022    & T  & Self & P  &       & Point  &     &    & \ding{51}\\
     \cmidrule(l){2-12} 
     & \multirow{4}{*}{CNN (\ref{sub:rep-cnn})}  
     & TF-C \citep{zhang2022self} & 2022  & T  & Self & W &   & - &    &     &  \ding{51} \\
      & & DCdetector \citep{yang2023dcdetector} & 2023  & ST  & Self & W &   & Point + Subseq &    &     &  \ding{51} \\ 
        & & CARLA \citep{darban2023carla} & 2023 & ST  & Self    & W     &     & Point + Subseq &      &      &  \ding{51} \\ 
         &  & DACAD \citep{darban2024dacad} & 2024    & ST  & Self        & W &    & Point + Subseq  &   &    \\                  
          \bottomrule
    \multirow{6}{*}{\rotatebox[origin=c]{90}{Hybrid}}          & \multirow{2}{*}{AE (\ref{sub:hy-ae})}          
        & CAE-M \citep{zhang2021unsupervised} & 2021    & ST      & Un        & W     &       & Subseq         &                     &                \\
     &  & NSIBF* \citep{feng2021time} & 2021      & T       & Un        & W     &                            & Subseq         &                            &                            \\ \cmidrule(l){2-12} 
     & \multirow{2}{*}{RNN (\ref{sub:hy-rnn})}    
         & TAnoGAN \citep{bashar2020tanogan} & 2020 & T       & Un        & W     &                            & Subseq         &                            &         \\ 
         &  & NSIBF* \citep{feng2021time} & 2021    & T       & Un        & W     &                            & Subseq         &                            &               \\                   \cmidrule(l){2-12}
    & \multirow{2}{*}{GNN (\ref{sub:hy-gnn})}                          & MTAD-GAT \citep{zhao2020multivariate} & 2020   & ST      & Self      & W     & \ding{51} & Subseq         &        \\ &   & FuSAGNet \citep{han2022learning} & 2022   & ST      & Semi      & W     &     & Subseq         &
          
          \\ \bottomrule
\end{tabular}
}
\\
\tiny \raggedright $^1$ A: \textit{Approach}, 
$^2$ MA: \textit{Main Architecture}, 
$^3$ T/S: \textit{Temporal/Spatial} | Values: [S:\textit{Spatial}, T:\textit{Temporal}, ST:\textit{Spatio-Temporal}], 
$^4$ Su/Un: \textit{Supervised/Unsupervised} | Values: [Su: \textit{Supervised}, Un: \textit{Unsupervised}, Semi: \textit{Semi-supervised}, Self: \textit{Self-supervised}], 
$^5$ Input: \textit{P: point / W: window},
$^6$ Int: \textit{Interpretability}, $^7$ P/S: \textit{Point/Sub-sequence}, $^8$ Stc: \textit{Stochastic}, $^9$ Inc: \textit{Incremental}, $^10$ US: Univarite support, $^*$ \textit{Models with more than one main architecture.}, $"-"$ \textit{indicates a feature is not defined or mentioned.}
\end{table}

Moreover, the deep model processes the input in a step-by-step or end-to-end fashion (see Fig. \ref{fig:components}). In the first category (step-by-step), there is a learning module followed by an anomaly scoring module. It is possible to combine the two modules in the second category to learn anomaly scores using neural networks as an end-to-end process. An output of these models may be anomaly scores or binary labels for inputs. Contrary to algorithms whose objective is to improve representations, DevNet \citep{pang2019deep}, for example, introduces deviation networks to detect anomalies by leveraging a few labelled anomalies to achieve end-to-end learning for optimizing anomaly scores. 
End-to-end models in anomaly detection are designed to directly output the final classification of data points or subsequences as normal or anomalous, which includes the explicit labelling of these points. In contrast, step-by-step models typically generate intermediate outputs at each stage of the analysis, such as anomaly scores for each subsequence or point. These scores then require additional post-processing, such as thresholding, to determine if an input is anomalous. Common methods for establishing these thresholds include Nonparametric Dynamic Thresholding (NDT) \citep{hundman2018detecting} and Peaks-Over-Threshold (POT) \citep{siffer2017anomaly}, which help convert scores into final labels.

An anomaly score is mostly defined based on a loss function. In most of the reconstruction-based approaches, reconstruction probability is used, and in forecasting-based approaches, the prediction error is used to define an anomaly score. An anomaly score indicates the degree of an anomaly in each data point. Anomaly detection can be accomplished by ranking data points according to anomaly scores ($A_S$) and a decision score based on a $threshold$ value:
\begin{equation}
    | A_S | > threshold
\end{equation}

Evaluation metrics that are used in these papers are introduced in Appendix \ref{sec:eval_metric}

\subsection{Forecasting-Based Models} \label{sec:FM}
The forecasting-based approach uses a learned model to predict a point or subsequence based on a point or a recent window. In order to determine how anomalous the incoming values are, the predicted values are compared to their actual values and their deviations are considered as anomalous values. Most forecasting methods use a sliding window to forecast one point at a time. This is especially helpful in real-world anomaly detection situations where normal behaviour is in abundance, but anomalous behaviour is rare.

It is worth mentioning that some previous works such as \citep{ma2003online} use prediction error as a \textit{novelty} quantification rather than an anomaly score.
In the following subsections, different forecasting-based architectures are explained. 
\begin{figure}[t]
     \centering
     \begin{subfigure}[b]{0.3\textwidth}
         \centering
         \includegraphics[width=\textwidth]{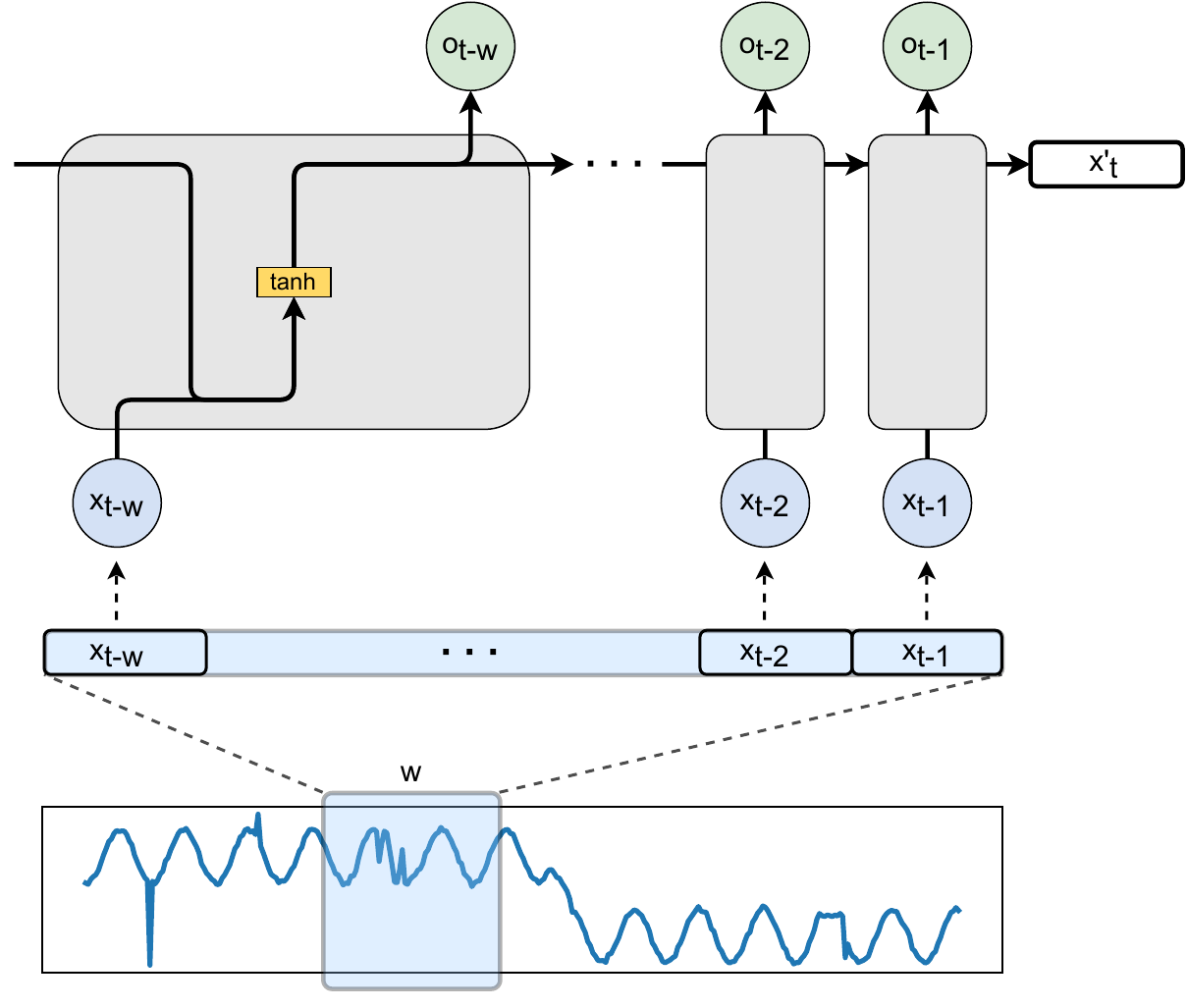}
         \caption{RNN}
         \label{fig:rnn:rnn}
     \end{subfigure}
     \hfill
     \begin{subfigure}[b]{0.3\textwidth}
         \centering
         \includegraphics[width=\textwidth]{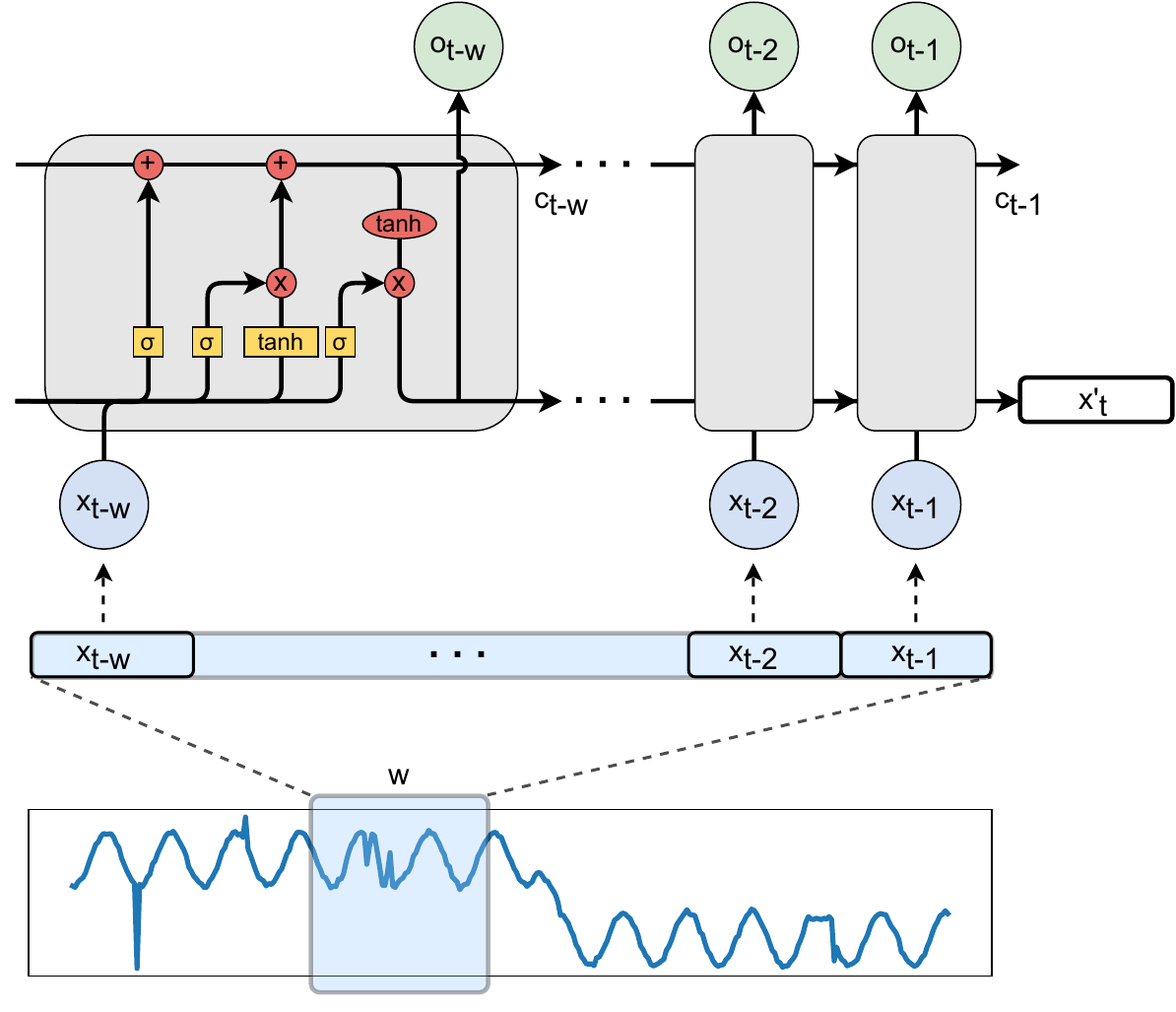}
         \caption{LSTM}
         \label{fig:rnn:lstm}
     \end{subfigure}
     \hfill
     \begin{subfigure}[b]{0.3\textwidth}
         \centering
         \includegraphics[width=\textwidth]{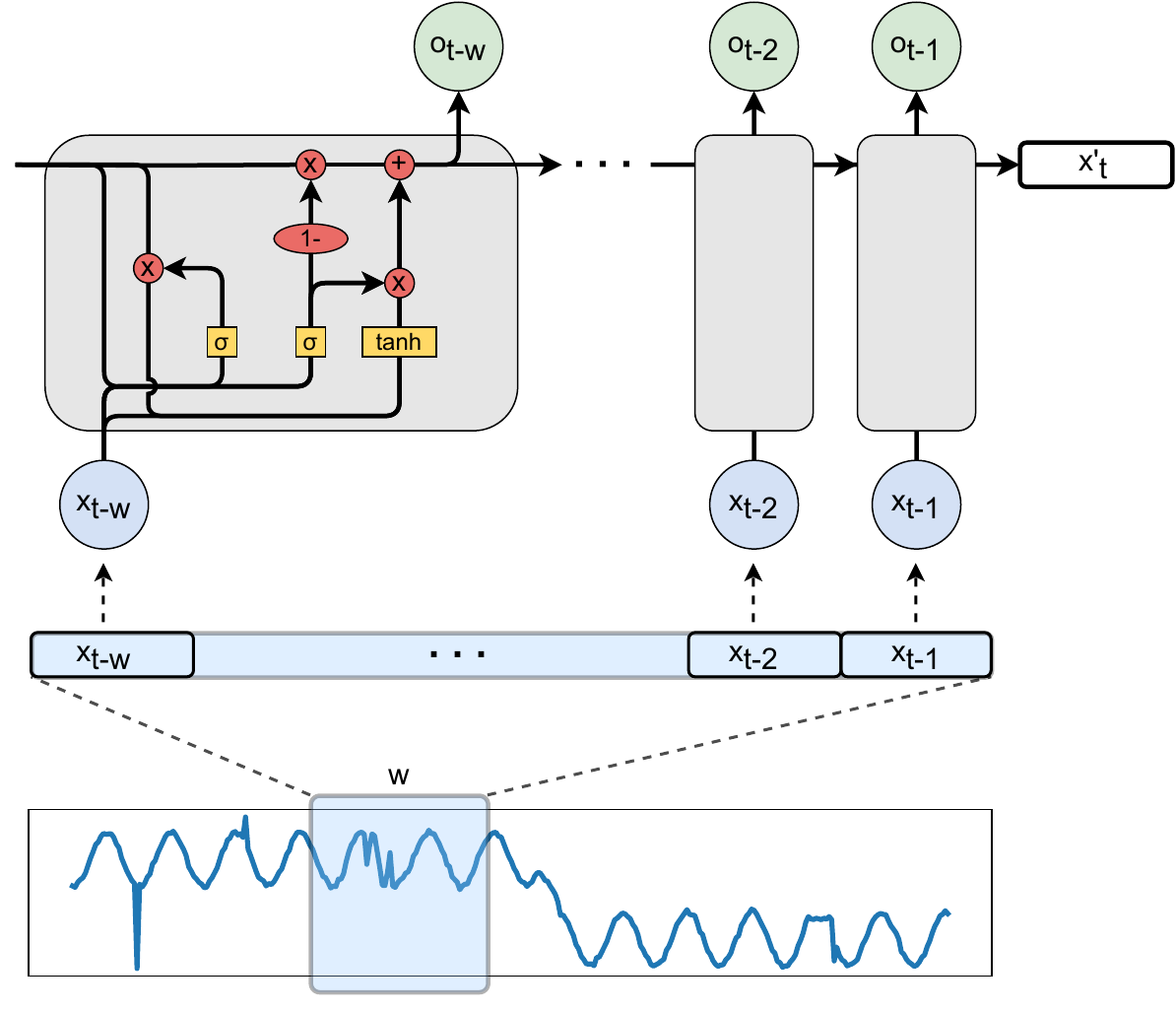}
         \caption{GRU}
         \label{fig:rnn:gru}
    \end{subfigure}
    \caption{An Overview of (a) Recurrent neural network (RNN), (b) Long short-term memory unit (LSTM), and (c) Gated recurrent unit (GRU). These models can predict $x_t'$ by capturing the temporal information of a window of $w$ samples prior to $x_t$ in the time series. Using the error $|x_t - x_t'|$, an anomaly score can be computed.}
    \label{fig:rnn}
    \vspace{-10pt}
\end{figure}
\subsubsection{Recurrent Neural Networks (RNN)} \label{sub:for-rnn}

RNNs have internal memory, allowing them to process variable-length input sequences and retain temporal dynamics \citep{tealab2018time, abiodun2018state}. An example of a simple RNN architecture is shown in Fig~\ref{fig:rnn:rnn}. Recurrent units take the points of the input window $X_{t-w:t-1}$ and forecast the next timestamp $x'_t$. The input sequence is processed iteratively, timestamp by timestamp. Given input $x_{t-1}$ to the recurrent unit $o_{t-2}$ and an activation function like $\tanh$, the output $x'_t$ is calculated as follows:

\begin{equation}
\begin{split}
    x'_t = \sigma(W_{x'}.o_{t-1} + b_{x'})\ , \\
    o_{t-1} = \tanh(W_o.x_{t-1} + U_o.o_{t-2} + b_h)
    \label{eq:RNN}
\end{split}
\end{equation}

where $W_{x'}$, $W_o$, $U_o$, and $b_h$ are the network parameters. The network learns long-term and short-term temporal dependencies using previous outputs as inputs.

LSTM networks extend RNNs with memory lasting thousands of steps \citep{hochreiter1997long}, enabling superior predictions through long-term dependencies. An LSTM unit, illustrated in Fig. \ref{fig:rnn:lstm}, comprises cells, input gates, output gates, and forget gates. The cell remembers values for variable time periods, while the gates control the flow of information.

In LSTM processing, the forget gate $f_{t-1}$ is calculated as:

\begin{equation}
    f_{t-1} = \sigma(W_f.x_{t-1} + U_f.o_{t-2})
\end{equation}
\begin{equation}
    i_{t-1} = \sigma(W_i.x_{t-1} + U_i.o_{t-2})
\end{equation}
\begin{equation}
    s_{t-1} = \sigma(W_s.x_{t-1} + U_s.o_{t-2})
\end{equation}

Next, the candidate cell state $\tilde{c_{t-1}}$ is updated as:

\begin{equation}
\begin{split}
    \tilde{c_{t-1}} = \tanh(W_c.x_{t-1} + U_c.o_{t-2})\ ,\\
    c_{t-1} = i_{t-1}.\tilde{c_{t-1}} + f_{t-1}.c_{t-2}
\end{split}
\end{equation}

Finally, the hidden state $o_{t-1}$ or output is:

\begin{equation}
   o_{t-1} = \tanh(c_{t-1}).s_{t-1}
\end{equation}

Where $W$ and $U$ are the parameters of the LSTM cell. $x'_t$ is finally calculated using Equation \ref{eq:RNN}.

Experience with LSTM has shown that stacking recurrent hidden layers with sigmoidal activation units effectively captures the structure of time series data, allowing for processing at different time scales compared to other deep learning architectures \citep{hermans2013training}. 
LSTM-AD \citep{malhotra2015long} possesses long-term memory capabilities and combines hierarchical recurrent layers to detect anomalies in UTS without using labelled data for training. This stacking helps learn higher-order temporal patterns without needing prior knowledge of their duration. The network predicts several future time steps to capture the sequence's temporal structure, resulting in multiple error values for each point in the sequence. These prediction errors are modelled as a multivariate Gaussian distribution to assess the likelihood of anomalies. LSTM-AD's results suggest that LSTM-based models are more effective than RNN-based models, especially when it's unclear whether normal behaviour involves long-term dependencies.

As opposed to the stacked LSTM used in LSTM-AD, \citet{bontemps2016collective} propose a simpler LSTM RNN model for collective anomaly detection based on its predictive abilities for UTS. First, an LSTM RNN is trained with normal time series data to make predictions, considering both current states and historical data. By introducing a circular array, the model detects collective anomalies by identifying prediction errors that exceed a certain threshold within a sequence.

Motivated by promising results in LSTM models for UTS anomaly detection, a number of methods attempt to detect anomalies in MTS based on LSTM architectures. 
In DeepLSTM \citep{chauhan2015anomaly}, stacked LSTM recurrent networks are trained on normal time series data. The prediction errors are then fitted to a multivariate Gaussian using maximum likelihood estimation. This model predicts both normal and anomalous data, recording the Probability Density Function (PDF) values of the errors. This approach has the advantage of not requiring preprocessing, and it works directly on raw time series.
LSTM-PRED \citep{goh2017anomaly} utilises three LSTM stacks with 100 hidden units each, processing data sequences of 100 seconds to learn temporal dependencies. Instead of setting thresholds for each sensor, it uses the Cumulative Sum (CUSUM) method to detect anomalies. CUSUM calculates the cumulative sum of the sequence predictions to identify small deviations, reducing false positives. It computes the positive and negative differences between predicted and actual values, setting Upper Control Limits (UCL) and Lower Control Limits (LCL) from the validation data to determine anomalies. Moreover, this model can pinpoint the specific sensor showing abnormal behaviour.

In all three above-mentioned models, LSTMs are stacked to improve prediction accuracy by analysing historical data from MTS; however, LSTM-NDT \citep{hundman2018detecting} combines various techniques. LSTM-NDT model introduces a technique that automatically adjusts thresholds for data changes, addressing issues like diversity and instability in evolving data. Another model, called LGMAD \citep{ding2019real}, enhances LSTM's structure for better anomaly detection in time series. Additionally, a method combines LSTM with a Gaussian Mixture Model (GMM) for detecting anomalies in both simple and complex systems, with a focus on assessing the system's health status through a health factor. This model can only be applied in low-dimensional applications. For high-dimensional data, it's suggested to use dimension reduction methods like PCA for effective anomaly detection \citep{huang2006network}.

\citet{ergen2019unsupervised} present LSTM-based anomaly detection algorithms in an unsupervised framework, as well as semi-supervised and fully supervised frameworks. To detect anomalies, it uses scoring functions implemented by One Class-SVM (OC-SVM) and Support Vector Data Description (SVDD) algorithms. In this framework, LSTM and OC-SVM (or SVDD) architecture parameters are jointly trained with well-defined objective functions, utilising two joint optimisation approaches. The gradient-based joint optimisation method uses revised OC-SVM and SVDD formulations, illustrating their convergence to the original formulations. As a result of the LSTM-based structure, methods are able to process data sequences of variable length. Aside from that, the model is effective at detecting anomalies in time series data without preprocessing. Moreover, since the approach is generic, the LSTM architecture in this model can be replaced by a GRU (gated recurrent neural networks) architecture \citep{chung2014empirical}. 

GRU was proposed by \citet{cho2014properties} in 2014, similar to LSTM but incorporating a more straightforward structure that leads to less computing time (see Fig. \ref{fig:rnn:gru}). Both LSTM and GRU use gated architectures to control information flow. However, GRU has gating units that inflate the information flow inside the unit without having any separate memory unit, unlike LSTM \citep{dey2017gate}. There is no output gate but an update gate and a reset gate. Fig. \ref{fig:rnn:gru} shows the GRU cell that integrates the new input with the previous memory using its reset gate. The update gate defines how much of the last memory to keep \citep{gulli2017deep}. The issue is that LSTMs and GRUs are limited in learning complex seasonal patterns in multi-seasonal time series. As more hidden layers are stacked and the backpropagation distance (through time) is increased, accuracy can be improved. However, training may be costly. 

In this regard, the AD-LTI model is a forecasting tool that combines a GRU network with a method called Prophet to learn seasonal time series data without needing labelled data. It starts by breaking down the time series to highlight seasonal trends, which are then specifically fed into the GRU network for more effective learning. When making predictions, the model considers both the overall trends and specific seasonal patterns like weekly and daily changes. However, since it uses past data that might include anomalies, the projections might not always be reliable. To address this, it introduces a new measure called Local Trend Inconsistency (LTI), which assesses the likelihood of anomalies by comparing recent predictions against the probability of them being normal, overcoming the issue that there might be anomalous frames in history.

Traditional one-class classifiers are developed for fixed-dimension data and struggle with capturing temporal dependencies in time series data \citep{ruff2018deep}. A recent model, called THOC \citep{shen2020timeseries}, addresses this by using a complex network that includes a multilayer dilated RNN \citep{chang2017dilated} and hierarchical SVDD \citep{tax2004support}. This setup allows it to capture detailed temporal features at multiple scales (resolution) and efficiently recognise complex patterns in time series data. It improves upon older models by using information from various layers, not just the simplest features, and it detects anomalies by comparing current data against its normal pattern representation. In spite of the accomplishments of RNNs, they still face challenges in processing very long sequences due to their fixed window size.

\subsubsection{Convolutional Neural Networks (CNN)} \label{sub:for-cnn}
\begin{figure}[t]
  \centering
  \includegraphics[width=0.6\linewidth]{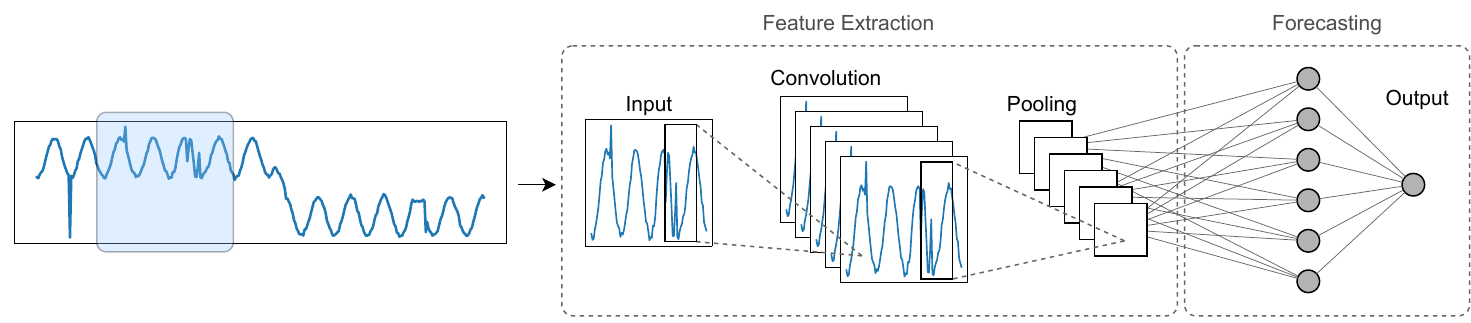}
  \caption{Structure of a Convolutional Neural Network (CNN) predicting the next values of an input time series based on a previous data window. Time series dependency dictates that predictions rely solely on previously observed inputs.}
  \label{fig:cnn}
  \vspace{-10pt}
\end{figure}

Convolutional Neural Networks (CNNs) are adaptations of multilayer perceptrons designed to identify hierarchical patterns in data. These networks employ convolutional, pooling, and fully connected layers, as depicted in Fig. \ref{fig:cnn}. Convolutional layers utilise a set of learnable filters that are applied across the entire input to produce 2D activation maps through dot products. Pooling layers summarise these outputs statistically.

The CNN-based DeepAnt model \citep{munir2018deepant} efficiently detects small deviations in time series patterns with minimal training data and can handle data contamination under 5\% in an unsupervised setup. DeepAnt is applicable to both UTS and MTS and detects various anomaly types, including point, contextual anomalies, and discords.

Despite their effectiveness, traditional CNNs struggle with sequential data due to their inherent design. This limitation has been addressed by the development of Temporal Convolutional Networks (TCN) \citep{bai2018empirical}, which use dilated convolutions to accommodate time series data. TCNs ensure that outputs are the same length as inputs without future data leakage. This is achieved using a 1D fully convolutional network and dilated convolutions, ensuring all computations for a timestamp \( t \) use only historical data. The dilated convolution operation is defined as:
\begin{equation}
    x'(t) = (x \ast_{l} f)(t) = \sum_{i=0}^{k-1} f(i) \cdot x_{t-l \cdot i}
\end{equation}
where \( f \) is a filter of size \( k \), \( \ast_{l} \) denotes convolution with dilation factor \( l \), and \( x_{t-l \cdot i} \) represents past data points.

\citet{he2019temporal} use different methods to predict and detect anomalies in data over time. They use a TCN trained on normal data to forecast trends and calculate anomaly scores using multivariate Gaussian distribution fitted to prediction errors. It includes a skipping connection to blend multi-scale features, accommodating different pattern sizes. \citet{ren2019time} combines a Spectral Residual model, originally for visual saliency detection \citep{hou2007saliency}, with a CNN to enhance accuracy. This method, used by over 200 Microsoft teams, can rapidly detect anomalies in millions of time series per minute.
The TCN Autoencoder (TCN-AE), developed by \citet{thill2020time} (2020), modifies the standard AE by using CNNs instead of dense layers, making it more effective and adaptable. It uses two TCNs for encoding and decoding, with layers that respectively downsample and upsample data.

Many real-world scenarios produce quasi-periodic time series (QTS), like the patterns seen in ECGs (electrocardiograms). A new automated system for spotting anomalies in these QTS called AQADF \citep{liu2020anomaly}, uses a two-part method. First, it segments the QTS into consistent periods using an algorithm (TCQSA) that uses a hierarchical clustering technique and groups similar data points without needing manual help, even filtering out errors to make it more reliable. Second, it analyses these segments with an attention-based hybrid LSTM-CNN model (HALCM), which looks at both broad trends and detailed features in the data. Furthermore, HALCM is further enhanced by three attention mechanisms, allowing it to capture more precise details of the fluctuation patterns in QTS. Specifically, TAGs are embedded in LSTMs in order to fine-tune variations extracted from different parts of QTS. A feature attention mechanism and a location attention mechanism are embedded into a CNN in order to enhance the effects of key features extracted from QTSs.

TimesNet \citep{wu2023timesnet} is a versatile deep learning model designed for comprehensive time series analysis. It transforms 1D time series data into 2D tensors to effectively capture complex temporal patterns. By using a modular structure called TimesBlock, which incorporates a parameter-efficient inception block, TimesNet excels in a variety of tasks, including forecasting, classification, and anomaly detection. This innovative approach allows it to handle intricate variations in time series data, making it suitable for applications across different domains.

\subsubsection{Graph Neural Networks (GNN)} \label{sub:for-gnn}
\begin{figure}[t]
  \centering
  \includegraphics[width=0.7\linewidth]{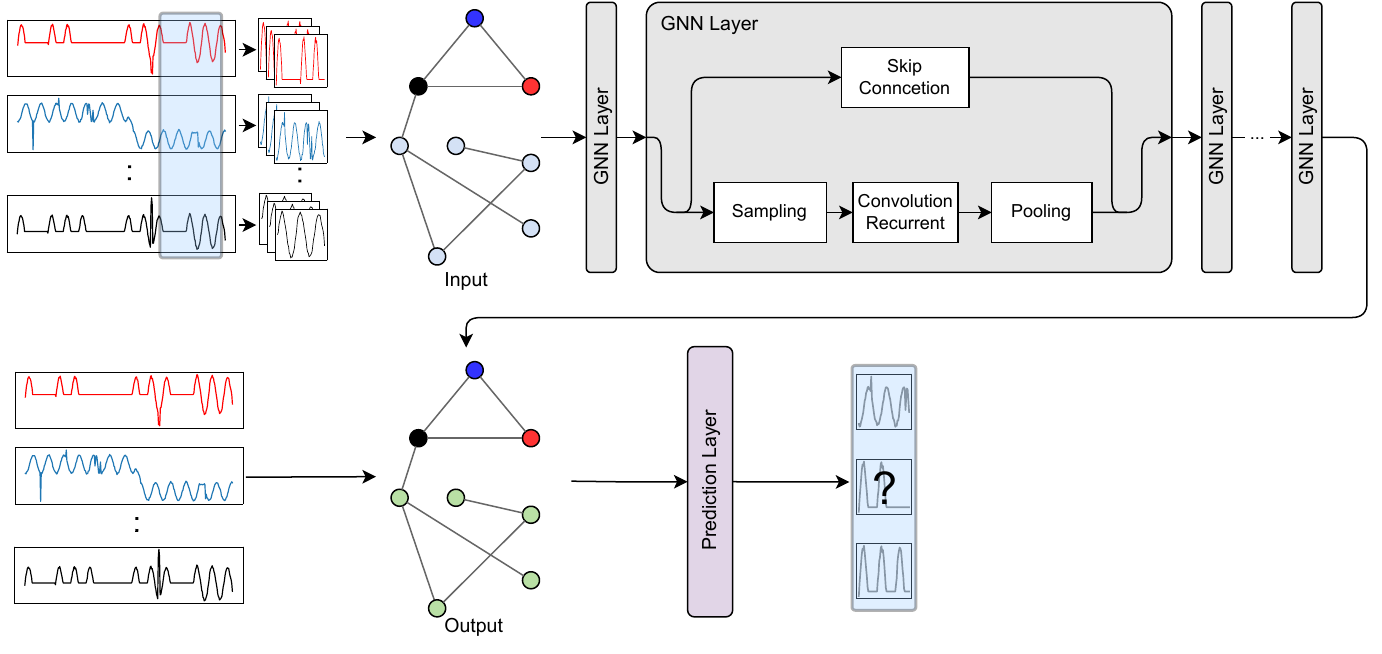}
  \caption{The basic structure of Graph Neural Network (GNN) for MTS anomaly detection that can learn the relationships (correlations) between metrics and predict the expected behaviour of time series.}
  \label{fig:gnn}
  \vspace{-10pt}
\end{figure}

In recent years, researchers have proposed extracting spatial information from MTS to form a graph structure, converting TSAD into a problem of detecting anomalies based on these graphs using GNNs.
As shown in Fig. \ref{fig:gnn}, GNNs use pairwise message passing, where graph nodes iteratively update their representations by exchanging information. In MTS anomaly detection, each dimension is a node in the graph, represented as $V = \{1, \ldots, d\}$. Edges \(E\) indicate correlations learned from MTS. For node \(u \in V\), the message passing layer outputs for iteration \(k+1\):

\begin{equation}
\begin{split}
    {h}_{u}^{k+1} = \text{UPDATE}^k (h_{u}^k, \ m_{N(u)}^k)\ , \\
    m_{N(u)}^k = \text{AGGREGATE}^k({h}_{i}^k, \forall i \in N(u))
\end{split}
\end{equation}

where \(h_u^k\) is the embedding for each node and \(N(u)\) is the neighbourhood of node \(u\). GNNs enhance MTS modelling by learning spatial structures \citep{scarselli2008graph}. Various GNN architectures exist, such as Graph Convolution Networks (GCN) \citep{kipf2016semi}, which aggregate one-step neighbours, and Graph Attention Networks (GAT) \citep{velivckovic2017graph}, which use attention functions to compute different weights for each neighbour.

Incorporating relationships between features is beneficial. \citet{deng2021graph} introduced GDN, a GNN attention-based model that captures sensor characteristics as nodes and their correlations as edges, predicting behaviour based on adjacent sensors. Anomaly detection framework GANF (Graph-Augmented Normalizing Flow) \citep{dai2022graph} augments normalizing flow with graph structure learning, detecting anomalies by identifying low-density instances. GANF represents time series as a Bayesian network, learning conditional densities with a graph-based dependency encoder and using graph adjacency matrix optimisation \citep{yu2019dag}.

In conclusion, extracting graph structures from time series and modelling them with GNNs enables the detection of spatial changes over time, representing a promising research direction.

\subsubsection{Hierarchical Temporal Memory (HTM)} \label{sub:for-htm}
\begin{figure}[t]
  \centering
  \begin{subfigure}[a]{\textwidth}
  \centering
    \includegraphics[width=0.6\linewidth]{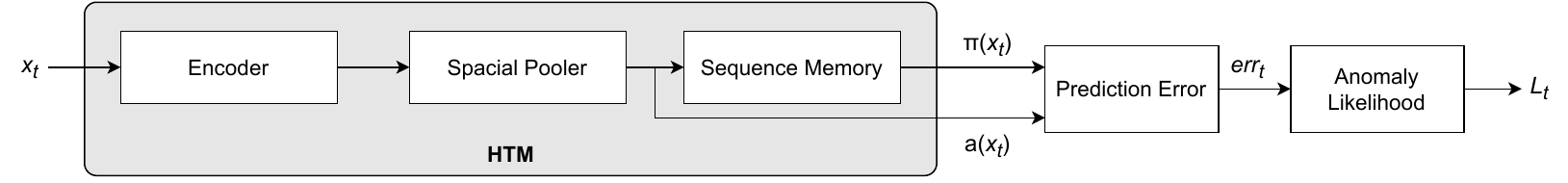}
    \caption{Components of anomaly detection using HTM}
    \label{fig:htmpipeline}
  \end{subfigure}
  \hfill\\
  \begin{subfigure}[b]{\textwidth}
    \centering
    \includegraphics[width=0.5\linewidth]{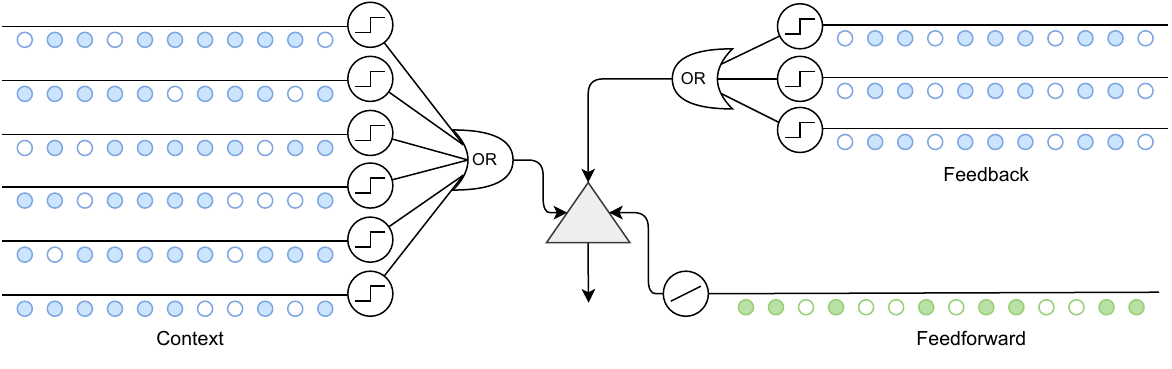}
    \caption{Structure of HTM cell}
    \label{fig:htm}
  \end{subfigure}
  \caption{(a) Components of an HTM-based (Hierarchical Temporal Memory) anomaly detection system calculating prediction error and anomaly likelihood. (b) An HTM cell internal structure. Dendrites act as detectors with synapses. Context dendrites receive lateral input from other neurons. Sufficient lateral activity puts the cell in a predicted state.}
  \vspace{-10pt}
\end{figure}

Hierarchical Temporal Memory (HTM) mimics the hierarchical processing of the neocortex for anomaly detection \citep{george2008brain}. Fig. \ref{fig:htmpipeline} shows the typical components of the HTM. The input $x_t$ is encoded and then processed through sparse spatial pooling \citep{cui2017htm}, resulting in $a(x_t)$, a sparse binary vector. Sequence memory models temporal patterns in $a(x_t)$ and returns a sparse vector prediction $\pi(x_t)$. The prediction error is defined as:

\begin{equation}
    err_t = 1 - \dfrac{\pi(x_{t-1}) \cdot a(x_t)}{|a(x_t)|}
\end{equation}

where $|a(x_t)|$ is the number of 1s in $a(x_t)$. Anomaly likelihood, based on the model's prediction history and error distribution, indicates whether the current state is anomalous.

HTM neurons are organised in columns within a layer (Fig. \ref{fig:htm}). Multiple regions exist within each hierarchical level, with fewer regions at higher levels combining patterns from lower levels to recognise more complex patterns. Sensory data enters lower-level regions during learning and generates patterns for higher levels. HTM is robust to noise, has high capacity, and can learn multiple patterns simultaneously. It recognises and memorises frequent spatial input patterns and identifies sequences likely to occur in succession.

Numenta HTM \citep{ahmad2017unsupervised} detects temporal anomalies of UTS in predictable and noisy environments. It effectively handles extremely noisy data, adapts continuously to changes, and can identify small anomalies without false alarms. Multi-HTM \citep{wu2018hierarchical} learns context over time, making it noise-tolerant and capable of real-time predictions for various anomaly detection challenges, so it can be used \mycomment{as} an adaptive model. In particular, it is used for univariate problems and applied efficiently to MTS.
RADM \citep{ding2018multivariate} proposes a real-time, unsupervised framework for detecting anomalies in MTS by combining HTM with a naive Bayesian network. Initially, HTM efficiently identifies anomalies in UTS with excellent results in terms of detection and response times. Then, it pairs with a Bayesian network to improve MTS anomaly detection without needing to reduce data dimensions, catching anomalies missed in UTS analyses. Bayesian networks help refine observations due to their adaptability and ease in calculating probabilities.

\subsubsection{Transformers} \label{sub:for-tran}
\begin{figure}[t]
  \centering
  \includegraphics[width=0.75\linewidth]{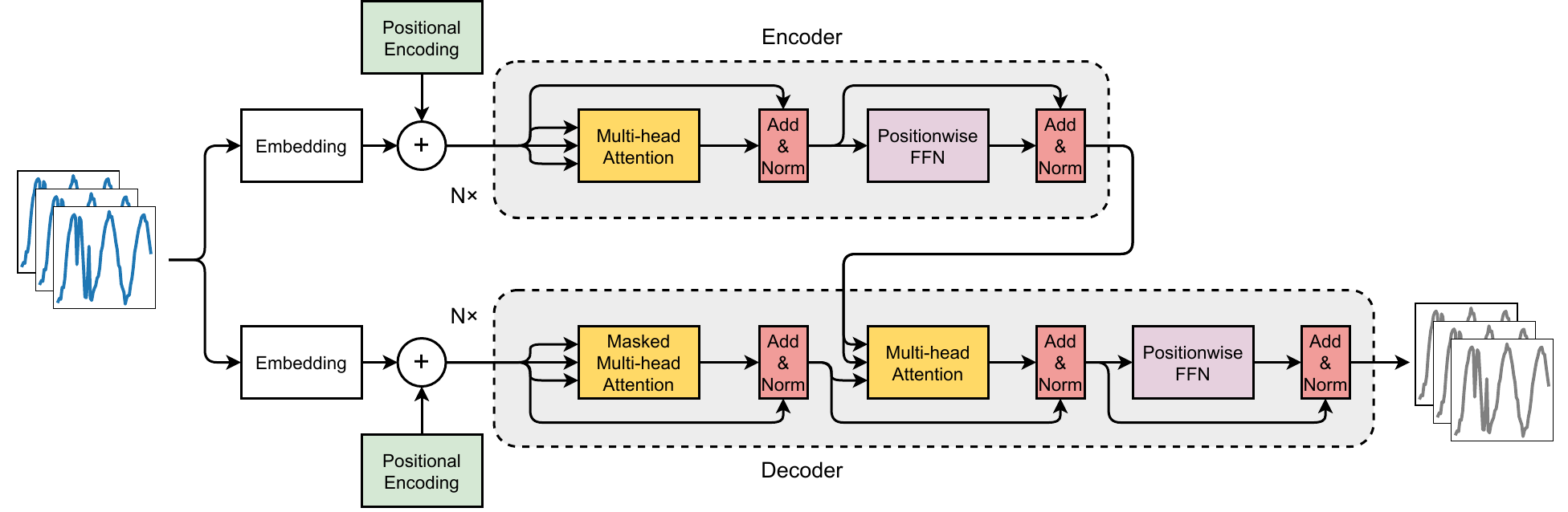}
  \caption{Transformer network structure for anomaly detection. The Transformer uses an encoder-decoder structure with multiple identical blocks. Each encoder block includes a multi-head self-attention module and a feedforward network. During decoding, cross-attention is added between the self-attention module and the feedforward network.}
  \label{fig:tran}
  \vspace{-10pt}
\end{figure}
Transformers \citep{vaswani2017attention} are deep learning models that weigh input data differently depending on the significance of different parts. In contrast to RNNs, transformers process the entire data simultaneously. Due to its architecture based solely on attention mechanisms, illustrated in Fig. \ref{fig:tran}, it can capture long-term dependencies while being computationally efficient. Recent studies utilise them to detect time series anomalies as they process sequential data for translation in text data. 

The original transformer architecture is encoder-decoder-based. An essential part of the transformer's functionality is its multi-head self-attention mechanism, stated in the following equation:
\begin{equation}
    Q,K,V = softmax({\frac {QK^{\mathrm {T} }}{\sqrt {d_{k}}}})V  
\end{equation}
where $Q$, $K$ and $V$ are defined as the matrices and $d_k$ is for normalisation of attention map.

A semantic correlation is identified in a long sequence, filtering out unimportant elements. Since transformers lack recurrence or convolution, they need positional encoding for token positions (i.e. relative or absolute positions). 
GTA \citep{chen2021learning} uses transformers for sequence modelling and a bidirectional graph to learn relationships among multiple IoT sensors. It introduces an Influence Propagation (IP) graph convolution for semi-supervised learning of sensor dependencies. To boost efficiency, each node's neighbourhood is constrained, and then graph convolution layers model information flow. As a next step, a multiscale dilated convolution and graph convolution are fused for hierarchical temporal context encoding. They use transformers for parallelism and contextual understanding and propose multi-branch attention to reduce attention complexity.
In another recent work, SAnD \citep{song2018attend} uses a transformer with stacked encoder-decoder structures, relying solely on attention mechanisms to model clinical time series. The architecture utilises self-attention to capture dependencies with multiple heads, positional encoding, and dense interpolation embedding for temporal order. It was also extended for multitask diagnoses.

\subsection{Reconstruction-Based Models} \label{sec:RM}
\begin{figure}[t]
     \centering
     \begin{subfigure}[b]{0.2\textwidth}
         \centering
         \includegraphics[width=\textwidth]{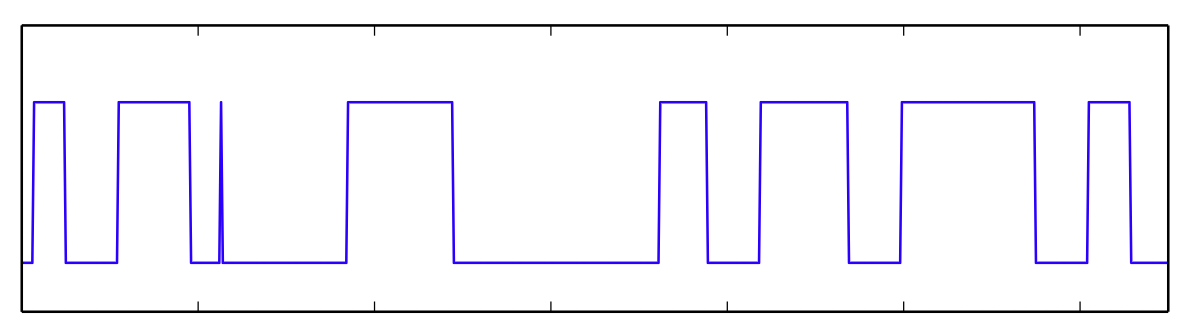}
         \caption{Predictable}
         \label{fig:ts:pred}
     \end{subfigure}
     \begin{subfigure}[b]{0.2\textwidth}
         \centering
         \includegraphics[width=\textwidth]{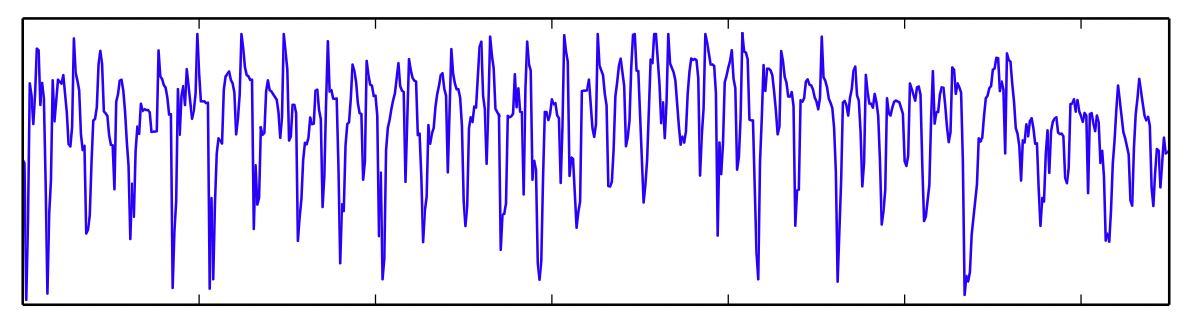}
         \caption{Unpredictable}
         \label{fig:ts:unpred}
     \end{subfigure}
    \caption{A time series may be unknown at any given moment or may change rapidly like (b), which illustrates sensor readings for manual control \citep{malhotra2016lstm}. Such a time series cannot be predicted in advance, making prediction-based anomaly detection ineffective.}
    \label{fig:ts}
    \vspace{-10pt}
\end{figure}
Many complex TSAD methods are designed around modelling the time series to predict future values, using prediction errors as indicators of anomalies. However, forecasting-based models often struggle with rapidly and continuously changing time series, as seen in Fig. \ref{fig:ts}, where the future states of a series may be unpredictable due to rapid changes or unknown elements \citep{golestani2014can}. In such cases, these models tend to generate increased prediction errors as the number of time points grows \citep{malhotra2015long}, limiting their utility primarily to very short-term predictions. For example, in financial markets, forecasting-based methods might predict only the next immediate step, which is insufficient in anticipating or mitigating a potential financial crisis.

In contrast, reconstruction-based models can offer more accurate anomaly detection because they have access to current time series data, which is not available to forecasting-based models. This access allows them to effectively reconstruct a complete scenario and identify deviations. While these models might cause some delay in detection, they are preferred when high accuracy is paramount, and some delay is acceptable. Thus, reconstruction-based models are better suited for applications where precision is critical, even if it results in a minor delay in response.

Models for normal behaviour are constructed by encoding subsequences of normal training data in latent spaces (low dimensions). Model inputs are sliding windows (see Section \ref{sec:DAD}) that provide the temporal context. We presume that the anomalous subsequences are less likely to be reconstructed compared to normal subsequences in the test phase since anomalies are rare. As a result, anomalies are detected by reconstructing a point/sliding window from test data and comparing them to the actual values, which is called reconstruction error. In some models, the detection of anomalies is triggered when the reconstruction probability is below a specified threshold since anomalous points/subsequences have a low reconstruction probability.

\subsubsection{Autoencoder (AE)} \label{sub:rec-ae}
Autoencoders (AEs), also known as auto-associative neural networks \citep{kramer1991nonlinear}, are widely used in MTS anomaly detection for their nonlinear dimensionality reduction capabilities \citep{sakurada2014anomaly, zong2018deep}. Recent advancements in deep learning have focused on learning low-dimensional representations (encoding) using AEs \citep{hinton2006reducing, bhatia2021mstream}. 

\begin{figure}[t]
     \centering
     \begin{subfigure}[b]{0.6\textwidth}
         \centering
         \includegraphics[width=\textwidth]{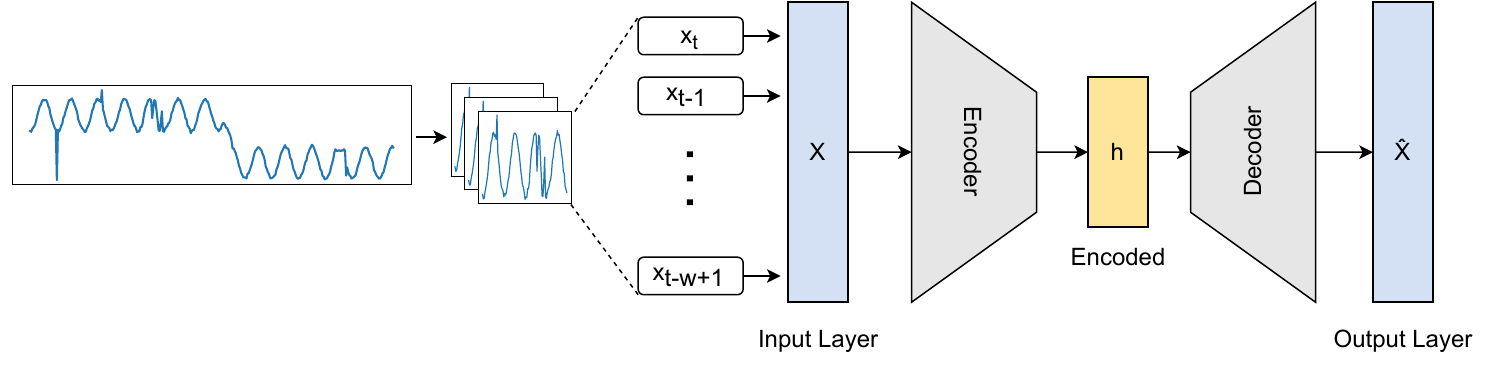}
         \caption{Auto-Encoder}
         \label{fig:ae:ae}
     \end{subfigure}
     \hfill
     \begin{subfigure}[b]{0.60\textwidth}
         \centering
         \includegraphics[width=\textwidth]{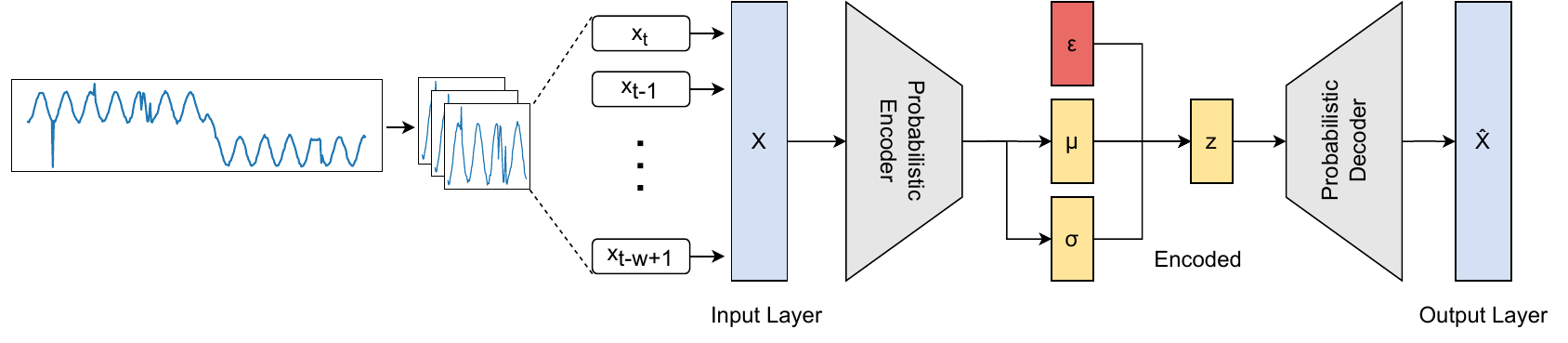}
         \caption{Variational Auto-Encoder}
         \label{fig:ae:vae}
     \end{subfigure}
    \caption{Structure of (a) Auto-Encoder that compresses an input window into a lower-dimensional representation ($h$) and then reconstructs the output $\hat{X}$ from this representation, and (b) Variational Auto-Encoder that its encoder compresses an input window of size $w$ into a latent distribution. The decoder uses sampled data from this distribution to produce $\hat{X}$, closely matching $X$.}
    \label{fig:ae}
    \vspace{-10pt}
\end{figure}

AEs consist of an encoder and a decoder (see Fig. \ref{fig:ae:ae}). The encoder converts input into a low-dimensional representation, and the decoder reconstructs the input from this representation. The goal is to achieve accurate reconstruction and minimise reconstruction error. This process is summarised as follows:

\begin{equation}
    Z_{t-w:t} = Enc(X_{t-w:t}, \phi), \quad \hat{X}_{t-w:t} = Dec(Z_{t-w:t}, \theta)
\end{equation}

where $X_{t-w:t}$ is a sliding window of input data, $x_t \in \mathbb{R}^d$, $Enc$ is the encoder with parameters $\phi$, and $Dec$ is the decoder with parameters $\theta$. $Z$ represents the latent space (encoded representation). The encoder and decoder parameters are optimised during training to minimise reconstruction error:

\begin{equation}
    (\phi^*, \theta^*) = \arg \min_{\phi, \theta} \text{Err}(X_{t-w:t}, Dec(Enc(X_{t-w:t}, \phi), \theta))
\end{equation}

To improve representation, techniques such as Sparse Autoencoder (SAE) \citep{ng2011sparse}, Denoising Autoencoder (DAE) \citep{vincent2008extracting}, and Convolutional Autoencoder (CAE) \citep{noh2015learning} are used. The anomaly score of a window in an AE-based model is defined based on the reconstruction error:

\begin{equation}
   AS_w = ||X_{t-w:t} - Dec(Enc(X_{t-w:t}, \phi), \theta)||^2
\end{equation}

There are several papers in this category in our study.  
\citet{sakurada2014anomaly} shows how AEs can be used for dimensionality reduction in MTS as a preprocessing step for anomaly detection. They treat each data sample at each time index as independent, disregarding the time sequence. Even though AEs already perform well without temporal information, they can be further boosted by providing current and past samples. The authors compare linear PCA, Denoising Autoencoders (DAEs), and kernel PCA, finding that AEs can detect anomalies that linear PCA  is incapable of detecting. DAEs further enhance AEs. Additionally, AEs avoid the complex computations of kernel PCA without losing quality in detection.
DAGMM (Deep Autoencoding Gaussian Mixture Model) \citep{zong2018deep} estimates the probability of MTS input samples using a Gaussian mixture prior to the latent space. It has two major components: a compression network for dimensionality reduction and an estimation network for anomaly detection using Gaussian Mixture Modelling to calculate anomaly scores in low-dimensional representations. However, DAGMM only considers spatial dependencies and lacks temporal information. The estimation network introduced a regularisation term that helps the compression network avoid local optima and reduce reconstruction errors through end-to-end training.

EncDec-AD \citep{malhotra2016lstm} model detects anomalies from unpredictable UTS by using the first principal component of the MTS. It can handle time series up to 500 points long but faces issues with error accumulation for longer sequences. \citep{kieu2019outlier} proposes two AEs ensemble frameworks based on sparsely connected RNNs: one with independent AEs and another with multiple AEs trained simultaneously, sharing features and using median reconstruction errors to detect outliers.
\citet{audibert2020usad} propose Unsupervised Anomaly Detection (USAD) using AEs in which adversarially trained AEs are utilised to amplify reconstruction errors in MTS, distinguishing anomalies and facilitating quick learning. The input to USAD for either training or testing is in a temporal order. \citet{goodge2020robustness} determine whether AEs are vulnerable to adversarial attacks in anomaly detection by analyzing the effects of various adversarial attacks. APAE (Approximate Projection Autoencoder) improves robustness against adversarial attacks by using gradient descent on latent representations and feature-weighting normalisation to account for variable reconstruction errors across features.

In MSCRED \citep{zhang2019deep}, attention-based ConvLSTM networks capture temporal trends, and a convolutional autoencoder (CAE) reconstructs a signature matrix, representing inter-sensor correlations instead of relying on the time series explicitly. The matrix length is 16, with a step interval of 5. An anomaly score is derived from the reconstruction error, aiding in anomaly detection, root cause identification, and anomaly duration interpretation.
In CAE-Ensemble \citep{campos2021unsupervised}, a convolutional sequence-to-sequence autoencoder captures temporal dependencies with high parallelism. Gated Linear Units (GLU) with convolution layers and attention capture local patterns, recognising recurring subsequences like periodicity. The ensemble combines outputs from diverse models based on CAEs and uses a parameter-transfer training strategy, which enhances accuracy and reduces training time and error. In order to ensure diversity, the objective function also considers the differences between basic models rather than simply assessing their accuracy.

RANSysCoders \citep{abdulaal2021practical} outlines a real-time anomaly detection system used by eBay. The authors propose an architecture with multiple encoders and decoders, using random feature selection and majority voting to infer and localise anomalies. The decoders set reconstruction bounds, functioning as bootstrapped AE for feature-bounds construction. The authors also recommend using spectral analysis of the latent space representation to extract priors for MTS synchronisation. Improved accuracy comes from feature synchronisation, bootstrapping, quantile loss, and majority voting. This method addresses issues with previous approaches, such as threshold identification, time window selection, downsampling, and inconsistent performance for large feature dimensions.

A novel Adaptive Memory Network with Self-supervised Learning (AMSL) \citep{zhang2022adaptive} is designed to increase the generalisation of unsupervised anomaly detection. AMSL uses an AE framework with convolutions for end-to-end training. It combines self-supervised learning and memory networks to handle limited normal data. The encoder maps the raw time series and its six transformations into a feature space. A multi-class classifier is then used to classify these features and improve generalisation. The features are also processed through global and local memory networks, which learn common and specific features. Finally, an adaptive fusion module merges these features into a new reconstruction representation.
Recently, ContextDA \citep{lai2023context} utilises deep reinforcement learning to optimise domain adaptation for TSAD. It frames context sampling as a Markov decision process, focusing on aligning windows from the source and target domains. The model uses a discriminator to align these domains without leveraging label information in the source domain, which may lead to ineffective alignment when anomaly classes differ. ContextDA addresses this by leveraging source labels, enhancing the alignment of normal samples and improving detection accuracy.

\subsubsection{Variational Autoencoder (VAE)} \label{sub:rec-vae}
Fig. \ref{fig:ae:vae} shows a typical configuration of the variational autoencoder (VAE), a directional probabilistic graph model which combines neural network autoencoders with mean-field variational Bayes \citep{kingma2013auto}. The VAE works similarly to AE, but instead of encoding inputs as single points, it encodes them as a distribution using inference network $q_{\phi}(Z_{t-w+1:t}|X_{t-w+1:t})$ where $\phi$ is its parameters. It represents a $d$ dimensional input $X_{t-w+1:t}$ to a latent representation $Z_{t-w+1:t}$ with a lower dimension $k<d$. A sampling layer takes a sample from a latent distribution and feeds it to the generative network $p_{\theta}(X_{t-w+1:t}|Z_{t-w+1:t})$ with parameters $\theta$, and its output is $g(Z_{t-w+1:t})$, reconstruction of the input. There are two components of the loss function, as stated in Equation (\ref{eq:vae}) that are minimised in a VAE: a reconstruction error that aims to improve the process of encoding and decoding and a regularisation factor, which aims to regularise the latent space by making the encoder's distribution as close to the preferred distribution as possible. 
\begin{equation}
\label{eq:vae}
    loss = ||X_{t-w+1:t} - g(Z_{t-w+1:t})||^2 + KL(N(\mu_x, \sigma_x),\ N(0,1))
\end{equation}
where $KL$ is the Kullback–Leibler divergence. By using regularised training, it avoids overfitting and ensures that the latent space is appropriate for a generative process.

LSTM-VAE \citep{park2018multimodal} represents a variation of the VAE that uses LSTM instead of a feed-forward network. This model is trained with a denoising autoencoding method for better representation. It detects anomalies when the log-likelihood of a data point is below a dynamic, state-based threshold to reduce false alarms. 
\citet{xu2018unsupervised} found that training on both normal and abnormal data is crucial for VAE anomaly detection. Their model, Donut, uses a VAE trained on shuffled data for unsupervised anomaly detection. Donut’s Modified ELBO, Missing Data Injection, and MCMC Imputation make it excellent at detecting anomalies in the seasonal KPI dataset. However, due to VAE's nonsequential nature and sliding window format, Donut struggles with temporal anomalies. Later on, Bagel \citep{li2018robust} is introduced to handle temporal anomalies robustly and unsupervised.
Instead of using VAE in Donut, Bagel employs conditional variational autoencoder (CVAE) \citep{lavin2015evaluating} and considers temporal information. VAE models the relationship between two random variables, $x$ and $z$. CVAE models the relationship between $x$ and $z$, conditioned on $y$, i.e., it models $p(x,z|y)$. 

STORNs \citep{solch2016variational}, or stochastic recurrent networks, use variational inference to model high-dimensional time series data. The algorithm is flexible and generic and doesn't need domain knowledge for structured time series. OmniAnomaly \citep{su2019robust} uses a VAE with stochastic RNNs for robust representations of multivariate data and planar normalizing flow for non-Gaussian latent space distributions. It detects anomalies based on reconstruction probability and uses POT for thresholding. InterFusion \citep{li2021multivariate} uses a hierarchical Variational Autoencoder (HVAE) with two stochastic latent variables for intermetric and temporal representations, along with a two-view embedding. To prevent overfitting anomalies in training data, InterFusion employs prefiltering temporal anomalies. The paper also introduces MCMC imputation, MTS for anomaly interpretation, and IPS for assessing results.

There are a few studies on anomaly detection in noisy time series data. Buzz \citep{chen2019unsupervised} uses an adversarial training method to capture patterns in univariate KPI with non-Gaussian noises and complex data distributions. This model links Bayesian networks with optimal transport theory using Wasserstein distance.
SISVAE (smoothness-inducing sequential VAE) \citep{li2020anomaly} detects point-level anomalies by smoothing before training a deep generative model using a Bayesian method. As a result, it benefits from the efficiency of classical optimisation models as well as the ability to model uncertainty with deep generative models. This model adjusts thresholds dynamically based on noise estimates, crucial for changing time series.
Other studies have used VAE for anomaly detection, assuming an unimodal Gaussian distribution as a prior. Existing studies have struggled to learn the complex distribution of time series due to its inherent multimodality. The GRU-based Gaussian Mixture VAE \citep{guo2018multidimensional} addresses this challenge of learning complex distributions by using GRU cells to discover time sequence correlations and represent multimodal data with a Gaussian Mixture.

In \citep{zhang2019velc}, a VAE with two extra modules is introduced: a Re-Encoder and a Latent Constraint network (VELC). The Re-Encoder generates new latent vectors, and this complex setup maximises the anomaly score (reconstruction error) in both the original and latent spaces to accurately model normal samples. The VELC network prevents the reconstruction of untrained anomalies, leading to latent variables similar to the training data, which helps distinguish normal from anomalous data.
The VAE and LSTM are integrated as a single component in PAD \citep{chen2021joint} to support unsupervised anomaly detection and robust prediction. The VAE minimises noise impact on predictions, while LSTMs help VAE capture long-term sequences. Spectral residuals (SR) \citep{hou2007saliency} are also used to improve performance by assigning weights to each subsequence, indicating their normality.

TopoMAD (topology-aware multivariate time series anomaly detector) \citep{he2020spatiotemporal} is an anomaly detector in cloud systems that uses GNN, LSTM, and VAE for spatiotemporal learning. It’s a stochastic seq2seq model that leverages topological information to identify anomalies using graph-based representations. The model replaces standard LSTM cells with graph neural networks (GCN and GAT) to capture spatial dependencies.
To improve anomaly detection, models like VAE-GAN \citep{niu2020lstm} use partially labelled data. This semi-supervised model integrates LSTMs into a VAE, training an encoder, generator, and discriminator simultaneously. The model distinguishes anomalies using both VAE reconstruction differences and discriminator results.

The recently developed Robust Deep State Space Model (RDSSM) \citep{li2022learning} is an unsupervised density reconstruction-based model for detecting anomalies in MTS. Unlike many current methods, RDSSM uses raw data that might contain anomalies during training. It incorporates two transition modules to handle temporal dependency and uncertainty. The emission model includes a heavy-tail distribution error buffer, allowing it to handle contaminated and unlabelled training data robustly. Using this generative model, they created a detection method that manages fluctuating noise over time. This model provides adaptive anomaly scores for probabilistic detection, outperforming many existing methods.

In \citep{wang2022variational}, a variational transformer is introduced for unsupervised anomaly detection in MTS. Instead of using a feature relationship graph, the model captures correlations through self-attention. The model's performance improves due to reduced dimensionality and sparse correlations. The transformer's positional encoding, or global temporal encoding, helps capture long-term dependencies. Multi-scale feature fusion allows the model to capture robust features from different time scales. The residual VAE module encodes hidden space using local features, and its residual structure improves the KL divergence and enhances model generation.

\subsubsection{Generative Adversarial Networks (GAN)} \label{sub:rec-gan}
\begin{figure}[t]
  \centering
  \includegraphics[width=0.6\linewidth]{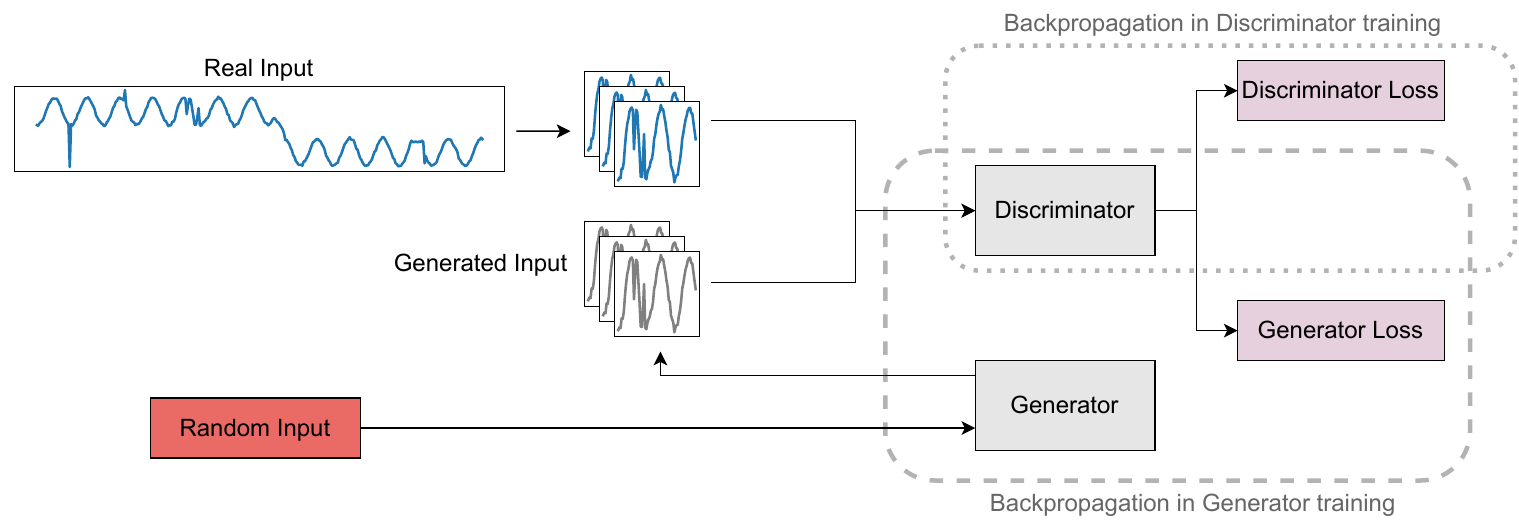}
  \caption{Overview of a Generative Adversarial Network (GAN) with two main components: generator and discriminator. The generator creates fake time series windows for the discriminator, which learns to distinguish between real and fake data. A combined anomaly score is calculated using both the trained discriminator and generator.}
  \label{fig:gan}
  \vspace{-10pt}
\end{figure}

A generative adversarial network (GAN) is an artificial intelligence algorithm designed for generative modelling based on game theory \citep{goodfellow2014generative}, \citep{goodfellow2014generative}. In generative models, training examples are explored, and the probability distribution that generated them is learned. In this way, GAN can generate more examples based on the estimated distribution, as illustrated in Fig. \ref{fig:gan}. Assume that we named the generator $G$ and the discriminator $D$. The generator and discriminator are trained using following minimax model: 
\begin{equation}
    \underset{G}min \  \underset{D}max  \ V(D, G) = \mathbb{E}_{x\sim p(X)} [log\ D(X_{t-w+1:t})] + \mathbb{E}_{z\sim p(Z)} [log(1-D(Z_{t-w+1:t}))]
\end{equation}
where $p(x)$ is the probability distribution of input data and $X_{t-w+1:t}$ is a sliding window from the training set, called real input in Fig.\ref{fig:gan}. Also, $p(z)$ is the prior probability distribution of the generated variable and $Z_{t-w+1:t}$ is a generated input window taken from a random space with the same window size.

In spite of the fact that GANs have been applied to a wide variety of purposes (mainly in research), they continue to involve unique challenges and research openings because they rely on game theory, which is distinct from most approaches to generative modelling. Generally, GAN-based models take into account the fact that adversarial learning makes the discriminator more sensitive to data outside the current dataset, making reconstructions of such data more challenging.
BeatGAN \citep{zhou2019beatgan} is able to regularise its reconstruction robustly because it utilises a combination of AEs and GANs \citep{goodfellow2014generative} in cases where labels are not available. Moreover, using the time series warping method improves detection accuracy by speed augmentation in training datasets and robust BeatGAN against variability involving time warping in time series data. Research shows that BeatGAN can detect anomalies accurately in both ECG and sensor data. 

However, training the GAN is usually difficult and requires a careful balance between the discriminator and generator \citep{kodali2017convergence}. A system based on adversarial training is not suitable for online use due to its instability and difficulty in convergence. 
With Adversarial Autoencoder Anomaly Detection Interpretation (DAEMON), anomalies are detected using adversarially generated time series. DAEMON’s training involves three steps. First, a one-dimensional CNN encodes MTS. Then, instead of decoding the hidden variable directly, a prior distribution is applied to the latent vector, and an adversarial strategy aligns the posterior distribution with the prior. This avoids inaccurate reconstructions of unseen patterns. Finally, a decoder reconstructs the time series, and another adversarial training step minimises differences between the original and reconstructed values.

MAD-GAN (Multivariate Anomaly Detection with GAN) \citep{li2019mad} is a GAN-based model that uses LSTM-RNN as both the generator and discriminator to capture temporal relationships in time series. It detects anomalies using reconstruction error and discrimination loss.
Furthermore, FGANomaly (Filter GAN) \citep{du2021gan} tackles overfitting in AE-based and GAN-based anomaly detection models by filtering out potential abnormal samples before training using pseudo-labels. The generator uses Adaptive Weight Loss, assigning weights based on reconstruction errors during training, allowing the model to focus on normal data and reduce overfitting.

\subsubsection{Transformers}\label{sub:rec-tran}
Anomaly Transformer \citep{xu2021anomaly} uses an attention mechanism to spot unusual patterns by simultaneously modelling prior and series associations for each timestamp. This makes rare anomalies more distinguishable. Anomalies are harder to connect with the entire series, while normal patterns connect more easily with nearby timestamps. Prior associations estimate a focus on nearby points using a Gaussian kernel, while series associations use self-attention weights from raw data. Along with reconstruction loss, a MINIMAX approach is used to enhance the difference between normal and abnormal association discrepancies.
TranAD \citep{tuli2022tranad} is another transformer-based model that has self-conditioning and adversarial training. As a result of its architecture, it is efficient for training and testing while preserving stability when dealing with huge input. When anomalies are subtle, transformer-based encoder-decoder networks may fail to detect them. However, TranAD's adversarial training amplifies reconstruction errors to fix this. Self-conditioning ensures robust feature retrieval, improving stability and generalisation.

\citet{li2021dct} present an unsupervised method called DCT-GAN, which uses a transformer to handle time series data, a GAN to reconstruct samples and spot anomalies, and dilated CNNs to capture temporal info from latent spaces. The model blends multiple transformer generators at different scales to enhance its generalisation and uses a weight-based mechanism to integrate generators, making it suitable for various anomalies. Additionally, MT-RVAE \citep{wang2022variational} significantly benefits from the transformer's sequence modelling and VAE capabilities that are categorised in both of these architectures. 

The Dual-TF \citep{nam2024breaking} is a framework for detecting anomalies in time series data by utilising both time and frequency information. It employs two parallel transformers to analyze data in these domains separately, then combines their losses to improve the detection of complex anomalies. This dual-domain approach helps accurately pinpoint both point-wise and subsequence-wise anomalies by overcoming the granularity discrepancies between time and frequency.

\subsection{Representation-Based Models}
\mycomment{Representation-based models aim to learn rich representations of input time series that can then be used in downstream tasks such as anomaly detection and classification. In other words, rather than using the time series in the raw input space for anomaly detection, the learned representations in the latent space are used for anomaly detection. By learning robust representations, these models can effectively handle the complexities of time series data, which often contains noise, non-stationarity, and seasonality. These models are particularly useful in scenarios where labelled data is scarce, as they can often learn useful representations in an unsupervised or self-supervised learning schemes. While time series representation learning has become a hot topic in the time series community and a number of attempts have been made in recent years, only limited work has targeted anomaly detection tasks, and this area of research is still largely unexplored. In the following subsections we surveyed representation-based TSAD models.}

\subsubsection{Transformers} \label{sub:rep-tran} 
TS2Vec \citep{yue2022ts2vec} utilises a hierarchical transformer architecture to capture contextual information at multiple scales, providing a universal representation learning approach \mycomment{using self-supervised contrastive learning} that defines anomaly detection problem as a downstream task across various time series datasets. In TS2Vec, positive pairs are representations at the same timestamp in two augmented contexts created by timestamp masking and random cropping, while negative samples are representations at different timestamps from the same series or from other series at the same timestamp within the batch. 

\subsubsection{Convolutional Neural Networks (CNN)}\label{sub:rep-cnn}
TF-C (Time-Frequency Consistency) model \citep{zhang2022self} is a self-supervised contrastive pre-training framework designed for time series data. By leveraging both time-based and frequency-based representations, the model ensures that these embeddings are consistent within a shared latent space through a novel consistency loss. Using 3-layer 1-D ResNets as the backbone for its time and frequency encoders, the model captures the temporal and spectral characteristics of time series. This architecture allows the TF-C model to learn generalisable representations that can be used for time series anomaly detection in downstream tasks. In TF-C, a positive pair consists slightly perturbed version of an original sample, while a negative pair includes different original samples or their perturbed versions.

DCdetector \citep{yang2023dcdetector} employs a deep CNN with a dual attention mechanism. This structure focuses on both spatial and temporal dimensions, using contrastive learning to enhance the separability of normal and anomalous patterns, making it adept at identifying subtle anomalies. 
In this model, a positive pair consists of representations from different views of the same time series, while it does not use negative samples and relies on the dual attention structure to distinguish anomalies by maximizing the representation discrepancy between normal and abnormal samples.

In contrast, CARLA \citep{darban2023carla} introduces a self-supervised contrastive representation learning approach using a two-phase framework. The first phase, called pretext, differentiates between anomaly-injected samples and original samples. In the second phase, self-supervised classification leverages information about the representations' neighbours to enhance anomaly detection by learning both normal behaviors and deviations indicating anomalies. 
In CARLA, positive pairs are selected from neighbours, while negative pairs are anomaly-injected samples. 
In the recent work, DACAD \citep{darban2024dacad} combines a TCN with unsupervised domain adaptation techniques in its contrastive learning framework. It introduces synthetic anomalies to improve learning and generalisation across different domains, using a structure that effectively identifies anomalies through enhanced feature extraction and domain-invariant learning. DACAD selects positive pairs and negative pairs similar to CARLA.

These models exemplify the advancement in using deep learning for TSAD, highlighting the shift towards models that not only detect but also understand the intricate patterns in time series data\mycomment{, which makes this area of research promising. Finally, while all the models in this category are based on self-supervised contrastive learning approaches, there is no work on self-prediction-based self-supervised approaches in the TSAD literature and this research direction is unexplored.}

\subsection{Hybrid Models}
These models integrate the strengths of different approaches to enhance time series anomaly detection. A forecasting-based model predicts the next timestamp, while a reconstruction-based model uses latent representations of the time series. Additionally, representation-based models learn comprehensive representations of the time series. By using a joint objective function, these combined models can be optimised simultaneously.

\subsubsection{Autoencoder (AE)} \label{sub:hy-ae}
By capturing spatiotemporal correlation in multisensor time series, the CAE-M (Deep Convolutional Autoencoding Memory network) \citep{zhang2021unsupervised} can model generalised patterns based on normalised data by undertaking reconstruction and prediction simultaneously. It uses a deep convolutional AE with a Maximum Mean Discrepancy (MMD) penalty to match a target distribution in low dimensions, which helps prevent overfitting due to noise or anomalies. To better capture temporal dependencies, it employs nonlinear bidirectional LSTMs with attention and linear autoregressive models.
Neural System Identification and Bayesian Filtering (NSIBF) \citep{feng2021time} is a new density-based TSAD approach for Cyber-Physical Security (CPS). It uses a neural network with a state-space model to track hidden state uncertainty over time, capturing CPS dynamics. In the detection phase, Bayesian filtering is applied to the state-space model to estimate the likelihood of observed values. This combination of neural networks and Bayesian filters allows NSIBF to accurately detect anomalies in noisy CPS sensor data.
\subsubsection{Recurrent Neural Networks (RNN)} \label{sub:hy-rnn}
With TAnoGan \citep{bashar2020tanogan}, they have developed a method that can detect anomalies in time series if a limited number of examples are provided. TAnoGan has been evaluated using 46 NAB time series datasets covering a range of topics. Experiments have shown that LSTM-based GANs can outperform LSTM-based GANs when challenged with time series data through adversarial training.
\subsubsection{Graph Neural Networks (GNN)} \label{sub:hy-gnn}
In \citep{zhao2020multivariate}, two parallel graph attention (GAT) layers are introduced for self-supervised multivariate TSAD. These layers identify connections between different time series and learn relationships between timestamps. The model combines forecasting and reconstruction approaches: the forecasting model predicts one point, while the reconstruction model learns a latent representation of the entire time series. The model can diagnose anomalous time series (interpretability).
FuSAGNet \citep{han2022learning} fused SAE reconstruction and GNN forecasting to find complex anomalies in multivariate data. It incorporates GDN \citep{deng2021graph} but embeds sensors in each process, followed by recurrent units to capture temporal patterns. By learning recurrent sensor embeddings and sparse latent representations, the GNN predicts expected behaviours during the testing phase.

\subsection{Model Selection Guidelines for Time Series Anomaly Detection}
This section provides a concise guideline for choosing \mycomment{a TSAD} method on specific characteristics of the data and the anomaly detection task \mycomment{at hand for practitioners to choose architectures that will provide the most accurate and efficient anomaly detection}.

\begin{itemize}
    \item \textbf{Multidimensional Data with Complex Dependencies:} GNNs are suitable for capturing both temporal and spatial dependencies in multivariate time series. They are particularly effective in scenarios such as IoT sensor networks and industrial systems where intricate interdependencies among \mycomment{dimensions} exist. GNN architectures \mycomment{such as} GCNs and GATs are \mycomment{suggested to be used in such settings}.
    \item \textbf{Sequential Data with Long-Term Temporal Dependencies:} LSTM and GRU are effective for applications requiring the modelling of long-term temporal dependencies. LSTM is commonly used in financial time series analysis, predictive maintenance, and healthcare monitoring. GRU, with its simpler structure, offers faster training times and is suitable for efficient temporal dependency modelling.
    \item \textbf{Large Datasets Requiring Scalability and Efficiency:} Transformers utilise self-attention mechanisms to efficiently model long-range dependencies, making them suitable for handling large-scale datasets \citep{khan2022transformers}, such as network traffic analysis. They are designed for robust anomaly detection by capturing complex temporal patterns, with models like the Anomaly Transformer \citep{xu2021anomaly} and TranAD \citep{tuli2022tranad} being notable examples.
    \item \textbf{Handling Noise in Anomaly Detection:} AEs and VAEs architectures are particularly adept at handling noise in the data, making them suitable for applications like network traffic, multivariate sensor data, and cyber-physical systems.
    \item \textbf{High-Frequency Data and Detailed Temporal Patterns:} CNNs are useful for capturing local temporal patterns in high-frequency data. They are particularly effective in detecting small deviations and subtle anomalies in data such as web traffic and real-time monitoring systems. TCNs extend CNNs by using dilated convolutions to capture long-term dependencies. As a result, they are suitable for applications where there exist long-range dependencies as well as local patterns \citep{bai2018empirical}.
    \item \textbf{Data with Evolving Patterns and Multimodal Distributions:} Combining the strengths of various architectures, hybrid models are designed to handle complex, high-dimensional time series data with evolving patterns like smart grid monitoring, industrial automation, and climate monitoring. These models, such as those integrating GNNs, VAEs, and LSTMs, are suitable for the mentioned applications.
    \item \textbf{Capturing Hierarchical and Multi-Scale Contexts:} HTM models are designed to capture hierarchical and multi-scale contexts in time series data. They are robust to noise and can learn multiple patterns simultaneously, making them suitable for applications involving complex temporal patterns and noisy data.
    \item \textbf{Generalisation Across Diverse Datasets:} Contrastive learning excels in scenarios requiring generalisation across diverse datasets by learning robust representations through positive and negative pairs. It effectively distinguishes normal from anomalous patterns in time series data, making it ideal for applications with varying conditions, such as industrial monitoring, network security, and healthcare diagnostics.
\end{itemize}

\section{Datasets} \label{sec:Dataset}
\begin{table}[t]
\caption{Public dataset and benchmarks used mostly for anomaly detection in time series. There are direct hyperlinks to their names in the first column.}
\label{tab:datasets}
\centering
\resizebox{0.85\columnwidth}{!}{
\begin{tabular}{@{}lllllll@{}}
\toprule
Dataset/Benchmark           & Real/Synth & MTS/UTS$^1$ & \# Samples$^2$       & \# Entities$^3$ & \# Dim$^4$ & Domain                     \\ \midrule
\href{https://archive.ics.uci.edu/ml/datasets/CalIt2+Building+People+Counts}{CalIt2} \citep{Dua:2019}      & Real       & MTS     & 10,080      & 2           & 2      & Urban events management    \\
\href{https://physionet.org/content/capslpdb/1.0.0/}{CAP} \citep{Terzano2001-ej} \citep{physionet}            & Real       & MTS     & 921,700,000 & 108         & 21     & Medical and health         \\
\href{https://www.unb.ca/cic/datasets/ids-2017.html}{CICIDS2017} \citep{sharafaldin2018toward}               & Real       & MTS     & 2,830,540   & 15          & 83     & Server machines monitoring \\
\href{https://www.openml.org/search?type=data\&sort=runs\&id=1597\&status=active}{Credit Card fraud detection} \citep{dal2015calibrating} & Real       & MTS     & 284,807     & 1           & 31     & Fraud detectcion           \\
\href{https://iair.mchtr.pw.edu.pl/Damadics}{DMDS} \citep{DMDS}                                                             & Real       & MTS     & 725,402     & 1           & 32     & Industrial Control Systems \\
\href{https://www.cs.ucr.edu/~eamonn/time_series_data_2018/}{Engine Dataset} \citep{UCRArchive2018}                          & Real       & MTS     & NA          & NA          & 12     & Industrial control systems \\
\href{https://github.com/exathlonbenchmark/exathlon}{Exathlon} \citep{jacob2020exathlon}                        & Real       & MTS     & 47,530      & 39          & 45     & Server machines monitoring \\
\href{https://zenodo.org/record/3884398\#.Y1NlUtJByRQ}{GECCO IoT} \citep{moritz_steffen_2018_3884398}                                              & Real       & MTS     & 139,566     & 1           & 9      & Internet of things (IoT)   \\
\href{https://www.kaggle.com/inIT-OWL/genesis-demonstrator-data-for-machine-learning}{Genesis} \citep{von2018anomaly}           & Real       & MTS     & 16,220      & 1           & 18     & Industrial control systems \\
\href{https://kas.pr/ics-research/dataset\_ghl\_1}{GHL} \citep{filonov2016multivariate}                           & Synth      & MTS     & 200,001     & 48          & 22     & Industrial control systems \\
\href{https://search.r-project.org/CRAN/refmans/fdm2id/html/ionosphere.html}{IOnsphere}  \citep{Dua:2019}                      & Real       & MTS     & 351         &             & 32     & Astronomical studies       \\
\href{https://kdd.ics.uci.edu/databases/kddcup99/kddcup99.html}{KDDCUP99} \citep{kdd99data}                                      & Real       & MTS     & 4,898,427   & 5           & 41     & Computer networks          \\
\href{https://archive.ics.uci.edu/ml/datasets/Kitsune+Network+Attack+Dataset}{Kitsune} \citep{Dua:2019}               & Real       & MTS     & 3,018,973   & 9           & 115    & Computer networks          \\
\href{https://github.com/QAZASDEDC/TopoMAD}{MBD} \citep{he2020spatiotemporal}                            & Real       & MTS     & 8,640       & 5           & 26     & Server machines monitoring \\
\href{https://archive.ics.uci.edu/ml/datasets/Metro+Interstate+Traffic+Volume}{Metro} \citep{Dua:2019}                         & Real       & MTS     & 48,204      & 1           & 5      & Urban events management    \\
\href{https://physionet.org/content/mitdb/1.0.0/}{MIT-BIH Arrhythmia (ECG)}  \citep{moody2001impact} \citep{physionet}                & Real       & MTS     & 28,600,000  & 48          & 2      & Medical and health         \\
\href{https://doi.org/10.13026/C2V30W}{MIT-BIH-SVDB}  \citep{greenwald1990improved} \citep{physionet}          & Real       & MTS     & 17,971,200  & 78          & 2      & Medical and health         \\
\href{https://github.com/QAZASDEDC/TopoMAD}{MMS} \citep{he2020spatiotemporal}                              & Real       & MTS     & 4,370       & 50          & 7      & Server machines monitoring \\
\href{https://github.com/khundman/telemanom}{MSL} \citep{hundman2018detecting}               & Real       & MTS     & 132,046     & 27          & 55     & Aerospace                  \\
\href{https://github.com/numenta/NAB}{NAB-realAdExchange} \citep{ahmad2017unsupervised}                                                     & Real       & MTS     & 9,616       & 3           & 2      & Business                   \\
\href{https://github.com/numenta/NAB}{NAB-realAWSCloudwatch} \citep{ahmad2017unsupervised}                                                  & Real       & MTS     & 67,644      & 1           & 17     & Server machines monitoring \\
\href{https://cs.fit.edu/~pkc/nasa/data/}{NASA Shuttle Valve Data} \citep{NASA_Shuttle_Valve}              & Real       & MTS     & 49,097      & 1           & 9      & Aerospace                  \\
\href{https://archive.ics.uci.edu/ml/datasets/URL+Reputation}{OPPORTUNITY} \citep{Dua:2019}                 & Real       & MTS     & 869,376     & 24          & 133    & Computer networks          \\
\href{https://github.com/eBay/RANSynCoders}{Pooled Server Metrics (PSM)} \citep{abdulaal2021practical}                  & Real       & MTS     & 132,480     & 1           & 24     & Server machines monitoring \\
\href{https://www.kaggle.com/datasets/nphantawee/pump-sensor-data}{PUMP} \citep{pump}                                       & Real       & MTS     & 220,302     & 1           & 44     & Industrial control systems \\
\href{https://github.com/khundman/telemanom}{SMAP} \citep{hundman2018detecting}                              & Real       & MTS     & 562,800     & 55          & 25     & Environmental management   \\
\href{https://github.com/NetManAIOps/OmniAnomaly/}{SMD} \citep{li2018robust}                    & Real       & MTS     & 1,416,825   & 28          & 38     & Server machines monitoring \\
\href{https://dataverse.harvard.edu/dataset.xhtml?persistentId=doi:10.7910/DVN/EBCFKM}{SWAN-SF}  \citep{DVN/EBCFKM_2020}                & Real       & MTS     & 355,330     & 5           & 51     & Astronomical studies       \\
\href{http://itrust.sutd.edu.sg/research/testbeds/secure-water-treatment-swat/}{SWaT} \citep{mathur2016swat}                         & Real       & MTS     & 946,719     & 1           & 51     & Industrial control systems \\
\href{https://itrust.sutd.edu.sg/testbeds/water-distribution-wadi/}{WADI} \citep{ahmed2017wadi}                                     & Real       & MTS     & 957,372     & 1           & 127    & Industrial control systems \\
\href{https://ride.citibikenyc.com/system-data}{NYC Bike} \citep{nycbike}                                                    & Real       & MTS/UTS & +25M         & NA          & NA     & Urban events management    \\
\href{https://www1.nyc.gov/site/tlc/about/tlc-trip-record-data.page}{NYC Taxi} \citep{nyctaxi}                                 & Real       & MTS/UTS & +200M        & NA          & NA     & Urban events management    \\
\href{https://www.cs.ucr.edu/~eamonn/time_series_data_2018/}{UCR} \citep{UCRArchive2018}                                      & Real/Synth & MTS/UTS & NA          & NA          & NA     & Multiple domains           \\
\href{https://archive.ics.uci.edu/ml/datasets/dodgers+loop+sensor}{Dodgers Loop Sensor Dataset} \citep{Dua:2019}                & Real       & UTS     & 50,400      & 1           & 1      & Urban events management    \\
\href{https://competition.aiops-challenge.com/home/competition/1484452272200032281}{KPI AIOPS} \citep{aiops_kpi}                                    & Real       & UTS     & 5,922,913   & 58          & 1      & Business                   \\
\href{https://github.com/MarkusThill/MGAB/.}{MGAB} \citep{markus_thill_2020_3760086}                            & Synth      & UTS     & 100,000     & 10          & 1      & Medical and health         \\
\href{https://doi.org/10.13026/C2KS3F}{MIT-BIH-LTDB} \citep{physionet}                                                          & Real       & UTS     & 67,944,954  & 7           & 1      & Medical and health         \\
\href{https://github.com/numenta/NAB}{NAB-artificialNoAnomaly}  \citep{ahmad2017unsupervised}                                               & Synth      & UTS     & 20,165      & 5           & 1      & -                          \\
\href{https://github.com/numenta/NAB}{NAB-artificialWithAnomaly} \citep{ahmad2017unsupervised}                                               & Synth      & UTS     & 24,192      & 6           & 1      & -                          \\
\href{https://github.com/numenta/NAB}{NAB-realKnownCause} \citep{ahmad2017unsupervised}                                                      & Real       & UTS     & 69,568      & 7           & 1      & Multiple domains           \\
\href{https://github.com/numenta/NAB}{NAB-realTraffic}  \citep{ahmad2017unsupervised}                                                        & Real       & UTS     & 15,662      & 7           & 1      & Urban events management    \\
\href{https://github.com/numenta/NAB}{NAB-realTweets} \citep{ahmad2017unsupervised}                                                          & Real       & UTS     & 158,511     & 10          & 1      & Business                   \\
\href{https://github.com/datamllab/tods/tree/benchmark/benchmark/synthetic}{NeurIPS-TS} \citep{lai2021revisiting}                        & Synth      & UTS     & NA          & 1           & 1      & -                          \\
\href{https://helios2.mi.parisdescartes.fr/~themisp/norma/}{NormA} \citep{boniol2021unsupervised}                                       & Real/Synth & UTS     & 1,756,524   & 21          & 1      & Multiple domains           \\
\href{https://www.cs.ucr.edu/~eamonn/time_series_data_2018/}{Power Demand Dataset} \citep{UCRArchive2018}                     & Real       & UTS     & 35,040      & 1           & 1      & Industrial control systems \\
\href{https://doi.org/10.5281/zenodo.2654726}{SensoreScope} \citep{guillermo_barrenetxea_2019_2654726}                                                    & Real       & UTS     & 621,874     & 23          & 1      & Internet of things (IoT)   \\
\href{https://www.cs.ucr.edu/~eamonn/time_series_data_2018/}{Space Shuttle Dataset} \citep{UCRArchive2018}                   & Real       & UTS     & 15,000      & 15          & 1      & Aerospace                  \\
\href{https://webscope.sandbox.yahoo.com/catalog.php?datatype=s\&did=70\&guccounter=1}{Yahoo} \citep{yahoo_ws5}             & Real/Synth & UTS     & 572,966     & 367         & 1      & Multiple domains           \\ \bottomrule
\end{tabular}
}
\\
\tiny\raggedright $^1$ MTS/UTS: \textit{Multivariate/Univariate},
$^2$ $\#$ samples: \textit{total number of samples}, $^3$ $\#$ Entities: \textit{number of distinct time series}, $^4$ $\#$ Dim: \textit{number of dimensions in MTS}
\vspace{-10pt}
\end{table}

This section summarises datasets and benchmarks for TSAD, which provides a rich resource for researchers in TSAD. Some of these datasets are single-purpose datasets for anomaly detection, and some are general-purpose time series datasets that we can use in anomaly detection model evaluation with some assumptions or customisation. We can characterise each dataset or benchmark based on multiple aspects and their natural features. Here, we collect 48 well-known and/or highly-cited datasets examined by classic and state-of-the-art (SOTA) deep models for anomaly detection in time series. These datasets are characterised based on the below attributes:

\begin{itemize}
\item Nature of the data generation which can be real, synthetic or combined.
\item Number of entities, which means the number of independent time series inside each dataset.
\item Type of variety for each dataset or benchmark, which can be multivariate, univariate or a combination of both.
\item Number of dimensions, which is the number of features of an entity inside the dataset.
\item Total number of samples of all entities in the dataset.
\item The application domain of the dataset.
\end{itemize}

Note some datasets have been updated by their authors and contributors occasionally or regularly over time. We considered and reported the latest update of the datasets and their attributes.
Table \ref{tab:datasets} shows all 48 datasets with all mentioned attributes for each of them. It also includes hyperlinks to the primary source to download the latest version of the datasets.

Based on our exploration, the commonly used MTS datasets in SOTA TSAD models are MSL \citep{hundman2018detecting}, SMAP \citep{hundman2018detecting}, SMD \citep{li2018robust}, SWaT \citep{mathur2016swat}, PSM \citep{abdulaal2021practical}, and WADI \citep{ahmed2017wadi}. For UTS, the commonly used datasets are Yahoo \citep{yahoo_ws5}, KPI \citep{aiops_kpi}, NAB \citep{ahmad2017unsupervised}, and UCR \citep{UCRArchive2018}. These datasets are frequently used to benchmark and compare the performance of different TSAD models.

More detailed information about these datasets can be found on this Github repository: \href{https://github.com/zamanzadeh/ts-anomaly-benchmark}{https://github.com/zamanzadeh/ts-anomaly-benchmark}.

\section{Discussion and Conclusion} \label{sec:Discuss}
In spite of the numerous advances in time series anomaly detection, there are still major challenges in detecting several types of anomalies (as described in Section \ref{sub:anoms}). In contrast to the tasks relating to the majority (regular patterns), anomaly detection focuses on minority, unpredictable and unusual events, which bring about some challenges. The following are some challenges that have to be overcome in order to detect anomalies in time series data using deep learning models:
\begin{itemize}
    \item System behaviour in the real world is highly dynamic and influenced by the prevailing environmental conditions, rendering time series data inherently non-stationary with frequently changing data distributions. This non-stationary nature necessitates the adaptation of deep learning models through online or incremental training approaches, enabling them to update continuously and detect anomalies in real-time. Such methodologies are crucial as they allow models to remain effective in the face of evolving patterns and sudden shifts, thereby ensuring timely and accurate anomaly detection.
    \item The detection of anomalies in multivariate high-dimensional time series data presents a particular challenge as \mycomment{data can become sparse in high dimension and the model} requires simultaneous consideration of both temporal dependencies and relationships between dimensions.
    \item In the absence of labelled anomalies, unsupervised, semi-supervised or \mycomment{self-supervised} approaches are required. Because of this, a large number of normal instances are incorrectly identified as anomalies. Hence, one of the key challenges is to find a mechanism for minimising false positives and improve recall rates of detection. 
    \item Time series datasets can exhibit significant differences in noise existence, and noisy instances may be irregularly distributed. Thus, models are vulnerable, and their performance is compromised by noise in the input data.
    \item The use of anomaly detection for diagnostic purposes requires interpretability. Even so, anomaly detection research focuses primarily on detection precision, failing to address the issue of interpretability.
    \item In addition to being rarely addressed in the literature, anomalies that occur on a periodic basis make detection more challenging. A periodic subsequence anomaly is a subsequence that repeats over time \citep{rasheed2013framework}. The periodic subsequence anomaly detection technique, in contrast to point anomaly detection, can be adapted in areas like fraud detection to identify periodic anomalous transactions over time.
\end{itemize}
The main objective of this study was to explore and identify state-of-the-art deep learning models for TSAD, industrial applications, and datasets. In this regard, a variety of perspectives have been explored regarding the characteristics of time series, types of anomalies in time series, and the structure of deep learning models for TSAD. On the basis of these perspectives, 64 recent deep models were comprehensively discussed and categorised. Moreover, time series deep anomaly detection applications across multiple domains were discussed along with datasets commonly used in this area of research. In the future, active research efforts on time series deep anomaly detection are necessary to overcome the challenges we discussed in this survey.

\bibliographystyle{ACM-Reference-Format}
\bibliography{main}

\newpage
\appendix
\section{Evaluation Metrics for Time Series Anomaly Detection} \label{sec:eval_metric}
Evaluating TSAD models is crucial for determining their effectiveness, especially in scenarios where anomalies are rare and tend to occur in sequences. Table \ref{tab:metrics} presents the key metrics used to assess TSAD model performance, including Precision, Recall, F1 Score, F1$_{PA}$ Score, AU-PR (Area Under the Precision-Recall Curve), AU-ROC (Area Under the Receiver Operating Characteristic Curve), MTTD (Mean Time to Detect), Affliation \citep{huet2022local}, and VUS \citep{paparrizos2022volume}. Detailed guidelines on when to utilise each metric and how to interpret their values are provided in Table \ref{tab:metricsguide}.

\begin{table}[h]
\footnotesize
\centering
\caption{Evaluation Metrics for Time Series Anomaly Detection}
\renewcommand{\arraystretch}{1}
\begin{tabularx}{\textwidth}{lXl}
\toprule
\textbf{Metric Name} & \textbf{Definition} & \textbf{Formula*} \\ \hline

\textbf{Precision} & The proportion of true positive results among all positive results predicted by the model. In time series anomaly detection, it indicates the accuracy of the detected anomalies. & 
$\text{Precision} = \frac{TP}{TP + FP}$ \\ \hline

\textbf{Recall} & The proportion of true positive results among all actual positive cases. It measures the model's ability to detect all actual anomalies. & 
$\text{Recall} = \frac{TP}{TP + FN}$ \\ \hline

\textbf{F1 Score} & The harmonic mean of precision and recall, providing a balance between the two metrics. It is useful when both precision and recall are important. & 
$\text{F1} = 2 \cdot \frac{\text{Precision} \cdot \text{Recall}}{\text{Precision} + \text{Recall}}$
\\ \hline

\textbf{F1$_{PA}$ Score} & \textbf{F1$_{PA}$ score}
is a F1 score utilise a segment-based evaluation techniques named point adjustment (PA), which at least one point within that segment is detected as abnormal~\citep{xu2018unsupervised}. This method can overestimate the performance of TSAD models (as mentioned in~\citep{darban2023carla}). & 
$\text{F1}_{\text{PA}} = 2 \cdot \frac{\text{Precision}_{\text{PA}} \cdot \text{Recall}_{\text{PA}}}{\text{Precision}_{\text{PA}} + \text{Recall}_{\text{PA}}}$
\\ \hline

\textbf{PA\%K} & \textbf{F1}$_{PA}$ is mitigated by employing a \textbf{PA\%K} protocol~\citep{schlegel2019towards} by focusing on segments of data $w$. A segment is considered correctly detected as anomalous if at least \( K \% \) of its points are true positives ($TP_w$). & 
$Accuracy_{w} = 
\begin{cases} 
1 & \text{if } \frac{TP_w}{|w|} \geq \frac{K}{100} \\
0 & \text{otherwise}
\end{cases}$ \\ \hline

\textbf{AU-PR} & Area Under Precision-Recall Curve is a performance measurement for classification problems at various threshold ($t$) settings. It is particularly useful for imbalanced datasets. & 
$\text{AU-PR} = \int_{0}^{1} \text{Precision}(t) \frac{d(\text{Recall}(t))}{dt} \, dt$ \\ \hline

\textbf{AU-ROC} & Area Under Receiver Operating Characteristic Curve represents the ability of the model to distinguish between classes based on different thresholds ($t$). A higher AU-ROC indicates better model performance. & 
$\text{AU-ROC} = \int_{0}^{1} \text{Recall}(t) \frac{d(\text{FPR}(t))}{dt} \, dt$
\\ \hline

\textbf{MTTD} & Mean Time to Detect is the average time taken to detect an anomaly at time $T_{\text{detect}}$ after it occurs in time $T_{\text{true}}$. This metric evaluates the model's responsiveness. & 
$\text{MTTD} = \frac{1}{n} \sum_{i=1}^{n} (T_{\text{detect}} - T_{\text{true}})$
\\ \hline

\textbf{Affliation} & The affiliation metric assesses the degree of overlap between the detected anomalies (D) and the actual anomalies (A). It is designed to provide a more nuanced evaluation by considering both the precision and recall of the detected anomalies. &

$\text{Affiliation} = \frac{|D \cap A|}{|D \cup A|}$

\\ \hline

\textbf{VUS} & The Volume Under the Surface quantifies the volume between the true anomaly signal $y$ and the predicted anomaly signal $\hat{y}$ over time. It captures both the temporal and amplitude differences between the two signals, providing a holistic measure of the detection performance. & $\text{VUS} = \int_0^T |y_t - \hat{y}_t| \, dt$
\\ \hline

\end{tabularx}
\label{tab:metrics}
\small\raggedright * TP: True Positive, FP: False Positive, TN: True Negative, FN: False Negative, FPR: FP/(FP + TN)

\end{table}

\begin{table}[h]
\footnotesize
\centering
\caption{Guideline to Use and Assess Evaluation Metrics}
\begin{tabularx}{\textwidth}{lXX}
\toprule
\textbf{Metric Name} & \textbf{Value Explanation} & \textbf{When to Use} \\ \hline

\textbf{Precision} & Low precision indicates many false alarms (normal instances classified as anomalies). High precision indicates most detected anomalies are actual anomalies, implying few false alarms. & Use when it is crucial to minimise false alarms and ensure that detected anomalies are truly significant. \\ \hline

\textbf{Recall} & Low recall indicates many true anomalies are missed, leading to undetected critical events. High recall indicates most anomalies are detected, ensuring prompt action on critical events. & Use when it is critical to detect all anomalies, even if it means tolerating some false alarms. \\ \hline

\textbf{F1} & Low F1 score indicates poor balance between precision and recall, leading to either many missed anomalies \mycomment{and/or} many false alarms. High F1 score indicates a good balance, ensuring reliable anomaly detection with minimal misses and false alarms. & Use when a balance between precision and recall is needed to ensure reliable overall performance. \\ \hline

\textbf{F1$_{PA}$ Score} & Low FF1$_{PA}$ indicates difficulty in accurately identifying the exact points of anomalies. High F1$_{PA}$ indicates effective handling of slight deviations, ensuring precise anomaly detection. & Use when anomalies may not be precisely aligned, and slight deviations in detection points are acceptable. \\ \hline

\textbf{PA\%K} & Low PA\%K indicates that the model struggles to detect a sufficient portion of the anomalous segment. High PA\%K indicates effective detection of segments, ensuring that a significant portion of the segment is identified as anomalous. & Use when evaluating the model's performance in detecting segments of anomalies rather than individual points. \\ \hline

\textbf{AU-PR} & Low AU-PR indicates poor model performance, especially with imbalanced datasets. High AU-PR indicates strong performance, maintaining high precision and recall across thresholds. & Use when dealing with imbalanced datasets, where anomalies are rare compared to normal instances. \\ \hline

\textbf{AU-ROC} & Low AU-ROC indicates the model struggles to distinguish between normal and anomalous patterns. High AU-ROC indicates effective differentiation, providing reliable anomaly detection. & Use for a general assessment of the model's ability to distinguish between normal and anomalous instances. \\ \hline

\textbf{MTTD} & High MTTD indicates significant delays in detecting anomalies. Low MTTD indicates quick detection, allowing prompt responses to critical events. & Use when the speed of anomaly detection is critical, and prompt action is required. \\ \hline

\textbf{Affiliation} & high value of the affiliation metric indicates a strong overlap or alignment between the detected anomalies and the true anomalies in a time series. & Use when a comprehensive evaluation is required, or the focus is early detection. \\ \hline

\textbf{VUS} & A lower VUS value indicates better performance, as it means the predicted anomaly signal is closer to the true signal. & Use when a holistic and threshold-free evaluation of time series anomaly detection methods is required. \\ \hline

\end{tabularx}
\label{tab:metricsguide}
\end{table}

\section{Interpretability Metrics} These metrics collectively offer a way to assess the interpretability of anomaly detection systems, specifically their ability to identify and prioritise the most relevant factors or dimensions contributing to each detected anomaly.

\textbf{HitRate@P\%} is defined in \citep{su2019robust} from HitRate@K used in recommender systems, modified to evaluate the accuracy of interpreting anomalies at the segment level. HitRate@P\% assesses whether the true causes (relevant dimensions) of an anomaly are included within the top P\% of the identified causes by the algorithm.

\begin{equation}
HitRate@P\% = \frac{\text{Number of true causes in top P\%}}{\text{Total number of true causes}}
\end{equation}

\textbf{Interpretation Score (IPS)} \citep{li2021multivariate}
is adapted from the concept of HitRate@K, provides a precise measure of interpretative performance by quantifying the model's ability to pinpoint the most relevant factors contributing to each anomaly.
It is typically defined in a manner that reflects the proportion of correctly identified causes within the top-k ranked items or factors, adjusted for their ranking order:

\begin{equation}
\text{IPS} = \frac{1}{N} \sum_{i=1}^{N} \frac{\text{Number of true causes in top k for segment } i}{\text{Total number of true causes in segment } i}
\end{equation}

Where \( N \) is the number of segments analyzed, and the counts are taken from the top k causes identified by the model for each segment.

\textbf{RC-top-k (Relevant Causes top-k)} metric \citep{garg2022evaluation} measures the fraction of events for which at least one of the identified causes is among the top k causes identified by the model. This metric focuses on the model's ability to capture at least one relevant cause out of the potentially several contributing factors.

\begin{equation}
\text{RC-top-k} = \frac{\text{Number of events with at least one true cause in top k}}{\text{Total number of events}}
\end{equation}

HitRate@P\% rewards identifying all of the true causes while RC-top-\mycomment{k} rewards identifying at least one of the causes.

\textbf{Reconstructed Discounted Cumulative Gain (RDCG@P\%)} is an adaptation (defined by \citep{chen2021daemon}) of the Normalised Discounted Cumulative Gain (NDCG), a well-known metric in information retrieval used to evaluate ranking quality. For anomaly detection, RDCG@P\% measures the effectiveness of the model in identifying the most relevant dimensions (causes) of an anomaly, based on their ranking according to the reconstruction error. \mycomment{The higher the error, the more likely it is that the dimension contributes significantly to the anomaly.}

\begin{equation}
\text{RDCG@P\%} = \sum_{i=1}^{P} \frac{2^{r_i} - 1}{\log_2(i+1)}
\end{equation}

Where \( r_i \) is the relevance score of the dimension at position \( i \) in the ranking, up to the top P\% of ranked dimensions.

\section{Experimental Results} \label{sec:experimental} 
\begin{figure}[h]
    \centering
    \includegraphics[width=1\linewidth]{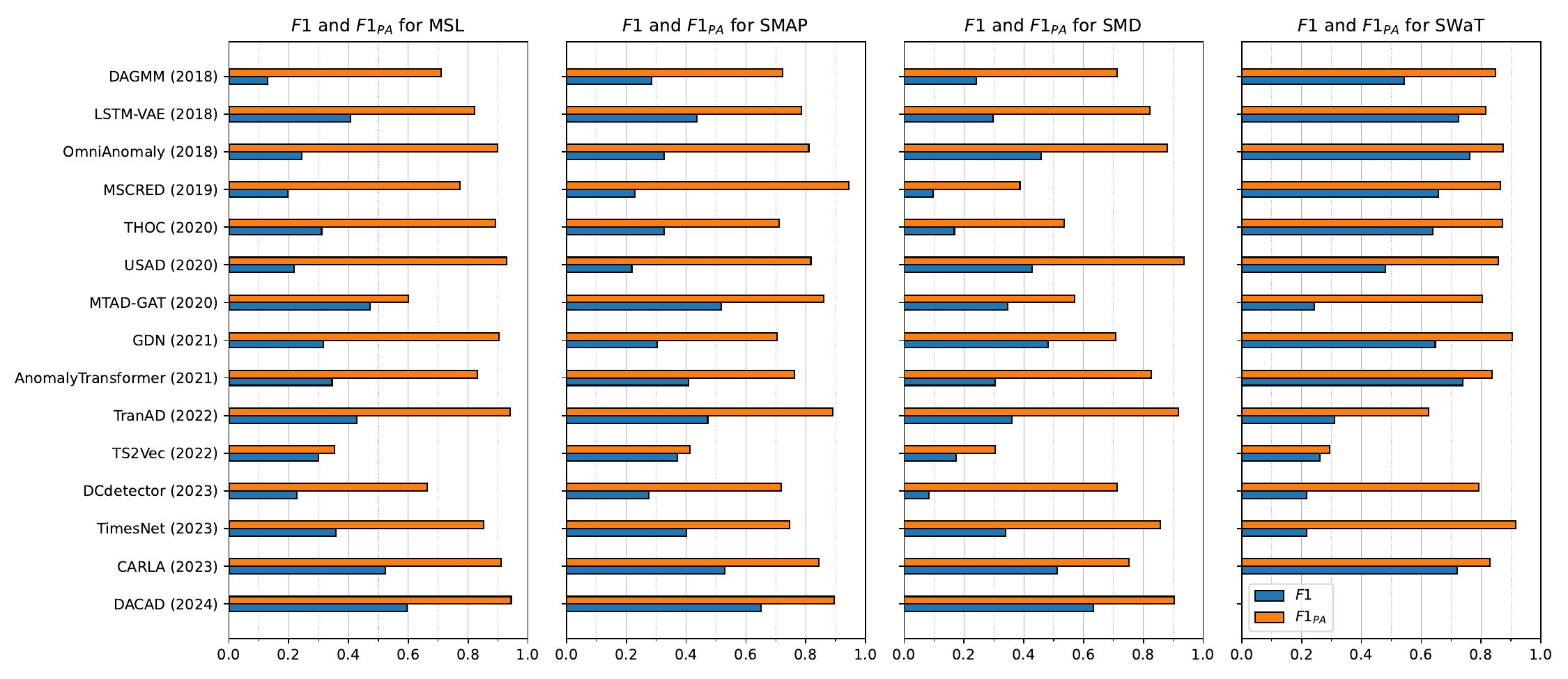}
    \caption{{$F1$} and {$F1_{PA}$} results for \mycomment{15 state-of-the-art TSAD} models on \mycomment{four most commonly used} MTS datasets.}
    \label{fig:experiments}
\end{figure}

The plots in Fig.~\ref{fig:experiments} provide a comparison of various TSAD models across the four different MTS datasets: MSL, SMAP, SMD, and SWaT datasets. Each model's performance is evaluated using two metrics: $F1$ score and $F1_{PA}$ score. Fig.~\ref{fig:experiments} illustrates that DACAD (2024) generally outperforms other models, especially on the MSL, SMAP, and SMD datasets, although it lacks results for the SWaT dataset since it cannot have results on it. CARLA (2023) and TimesNet (2023) also show strong performance across these datasets. In contrast, older models like DAGMM (2018), LSTM-VAE (2018), and OmniAnomaly (2018) generally exhibit lower scores compared to the more recent models. The performance improvement trend is evident with the newer models, which tend to achieve higher $F1$ and $F1_{PA}$ scores, indicating advancements in anomaly detection techniques over time.

\section{Application Areas of Deep Anomaly Detection in Time Series}\label{sec:APP}
An application typically generates data through a series of generating processes, which further reflect system operations or provide observational information about entities. The result of abnormal behaviour by the generating process is an anomaly. In other words, anomalies often reveal abnormal characteristics of the systems and entities used to generate data. By recognizing these unusual characteristics, we can gain useful insight from different applications. The following deep models are classified by the applications they are used for. 
\subsection{Computer Networks}
Intrusion detection for computer networks is becoming one of the most critical tasks for network administrators today. Traditional misuse detection strategies are unable to detect new and unknown intrusion types. In contrast, anomaly detection in network security aims to distinguish between malicious events and normal behaviour of network systems.

An essential part of defending a company's computer networks is the use of network intrusion detection systems (NIDS) to detect different security breaches. The feasibility and sustainability of contemporary networks are challenged by the need for increased human interaction and decreasing accuracy of detection. In \citep{javaid2016deep}, using deep learning techniques, they obtain a high-quality feature representation from unlabelled network traffic data and apply a supervised model on the KDD Cup 99 dataset \citep{tavallaee2009detailed}. Also, \citep{alrawashdeh2016toward}, a Restricted Boltzmann Machine (RBM) and a deep belief network are used for attack (anomaly) detection in KDD Cup 99. S-NDAE \citep{shone2018deep} is trained in an unsupervised manner to extract significant features from the dataset using unsupervised feature learning with nonsymmetric deep autoencoders (NDAEs).

With the rapid expansion of mobile data traffic and the number of connected devices and applications, it is necessary to establish a network management system capable of predicting and detecting anomalies effectively. A measure of latency in these networks is the round trip delay (RTT) between a probe and a central server that monitors radio availability. RCAD \citep{ahmed2022rcad} presents a distributed architecture for unsupervised detection of RTT anomalies, specifically increases in RTT. It employs the hierarchical temporal memory (HTM) algorithm to build a predictive model.

\subsection{Medicine and Health}
With the widespread adoption of electronic health records, there is an increased emphasis on predictive models that can effectively deal with clinical time series data. Some new approaches are intended to analyse physiological time series, identify potential risks of illness, and determine mitigation measures to take. \citep{wang2016research} uses several convolution layers to extract useful features from the input and then feeds them into a multivariate Gaussian distribution to detect anomalies. 

Electrocardiography (ECG) signals are frequently used to assess the health of the heart. A complex organ like the heart can cause many different arrhythmias. Thus, it would be very beneficial to adopt an anomaly detection approach for analysing ECG signals which are developed in \citep{kieu2019outlier}, \citep{zhou2019beatgan} and \citep{chauhan2015anomaly}.

Cardiovascular diseases (CVDs) are the leading cause of death in the world. Detecting abnormal heart rates can help doctors find patients' CVDs. Using a CNN, \citet{rubin2017recognizing} develop an automated recognition system for unusual heartbeats based on deep learning. In comparison to other popular deep models like CNN, RNNs are more effective at capturing the temporal characteristics of heartbeat sequences. A study on abnormal heartbeat detection using phonocardiography signals is presented in \citep{rubin2017recognizing}. It has been shown that RNNs are capable of producing promising results even in the presence of noise. Also, \citet{latif2018phonocardiographic} uses RNNs because of their ability to model sequential and temporal data, even in noisy environments, to detect abnormal heartbeats automatically. \citep{chen2020imbalanced} proposes a model using the classical echo state network (ESN) \citep{jaeger2007echo} trained on an imbalanced univariate heart rate dataset.

An epilepsy detection framework based on TCN, Gaussian mixture models and Bayesian inference called TCN-GMM \citep{liu2019anomaly} uses TCN to extract features from EEG time series.
Also, it is possible to treat Alzheimer's disease more effectively if the disease is detected early. A 2D-CNN randomised ensemble model is presented in \citep{lopez2020detection} that uses magnetoencephalography (MEG) synchronisation measures to detect early Alzheimer's disease symptoms.

\subsection{Internet Of Things (IoT)}
As part of the smart world, the Internet of Things (IoT) is playing an increasingly significant role in monitoring various industrial equipment used in power plants and handling emergency situations \citep{qiu2017eabs}. Analysing data anomalies can identify environmental circumstances that require human attention, uncover outliers when cleaning sensor data, or save computing resources by prefiltering undesirable portions of the data. 
Greenhouse \citep{lee2018greenhouse} applies a multi-step ahead predictive LSTM over high volumes of IoT time series. A semi-supervised hierarchical stacking TCN is presented in \citep{cheng2019hs}, which targets the detection of anomalies in smart homes' communication. Due to their use of offline learning, these approaches are not resistant to changes in input distribution.
In the Industrial Internet of Things (IIoT), massive amounts of data are generated, which are valuable for monitoring the status of the underlying equipment and boosting operational performance. An LSTM-based model is presented in \citep{zhang2018lstm} for analysis and forecasting of sensor data from IIoT devices to capture the time span surrounding the failures.
\citet{kim2018squeezed} perform unsupervised anomaly detection using real industrial IIoT time series, such as manufacturing CNC and UCI time series, using a Squeezed Convolutional Variational Autoencoder (SCVAE) deployed in an edge computing environment.

\subsection{Server Machines Monitoring and Maintenance}
Cloud systems have fueled the development of microservice architecture in the IT industry. A service failure in such an architecture can cause a series of failures, negatively impacting the customer experience and the company's revenue \citep{dragoni2017microservices}. Troubleshooting needs to be performed as soon as possible after an incident. For this reason, continuously monitoring online systems for any anomalies is essential. SLA-VAE \citep{huang2022semi} uses a semi-supervised VAE to identify anomalies in MTS in order to enhance robustness. Using active learning, a framework is designed that can learn and update a detection model online based on a small sample size of highly uncertain data. Cloud server data from two different types of game businesses are used for the experiments. For each cloud server, 11 monitored metrics, such as CPU usage, CPU load, disk usage, and memory usage, are adopted.

Detecting anomalies is essential in wireless sensor networks (WSNs) as it can reveal information about equipment faults and previously unknown events. \citet{luo2018distributed} introduces an AE-based model to solve anomaly detection problems in WSNs. The algorithm is designed to detect anomalies in sensors locally without requiring communication with other sensors or the cloud.

\subsection{Urban Events Management}
Traffic anomalies, such as traffic accidents and unexpected crowd gathering, may endanger public safety if not handled timely. However, traffic anomaly detection faces two main challenges. First, it is challenging to model traffic dynamics due to the complex spatiotemporal characteristics of data. Second, the criteria for traffic may vary with locations and times. 
\citet{zhang2019decomposition} outline a spatiotemporal decomposition framework, which is proposed for detecting urban anomalies. Spatial and temporal features are derived using a graph embedding algorithm to adapt to different locations and times.
\citep{deng2022graph} presents a traffic anomaly detection model based on a spatiotemporal graph convolutional adversarial network (STGAN). Spatiotemporal generators can be used to capture the spatiotemporal dependencies of traffic data.

In order to model dynamic multivariate data effectively, CHAT \citep{huang2021cross} is devised. In CHAT, the authors model the urban anomaly prediction problem based on hierarchical attention networks.
Uber uses an end-to-end neural network architecture for uncertainty estimation \citep{zhu2017deep}. To improve anomaly detection accuracy, the proposed uncertainty estimate is used to measure the uncertainty of special events (such as holidays).

One of the most challenging tasks in transportation is forecasting the speed of traffic. The use of a traffic prediction system prior to travel in urban areas can help drivers avoid potential congestion and reduce travel time. The aim of GTransformer \citep{lu2022graph} is to study how GNNs can be combined with attention mechanisms to improve traffic prediction accuracy. Also, TH-GAT \citep{huang2021temporal} is a temporal hierarchical graph attention network designed specifically for this purpose.

\subsection{Astronomical Studies}
As astronomical observations and data processing technology advance, an enormous amount of data is generated exponentially.``Light curve'' is generated using a series of processing steps on a star image. Studying light curves contributes to astronomy as a new method for detecting abnormal astronomical events \citep{li2001gamma}. In \citep{zhang2018time}, an LSTM  neural network is proposed for predicting light curves.

\subsection{Aerospace}
Due to the complexity and cost of spacecraft, failure to detect hazards during flight could lead to serious or even catastrophic destruction. In \citep{meng2019spacecraft}, a transformer-based model with two novel components is presented, namely, an attention mechanism that updates timestamps concurrently and a masking strategy that detects anomalies in advance. Testing was conducted on NASA telemetry datasets.

Monitoring and diagnosing the health of liquid rocket engines (LREs) is the most significant concern for spacecraft and vehicle safety, particularly for human launch. Failure of the engine will result directly in the failure of the space launch, resulting in irreparable losses. To achieve reliable and automatic anomaly detection for large equipment such as LREs and multisource data, \citet{feng2022unsupervised} suggest using a multimodal unsupervised method with missing sources. 

\subsection{Natural Disaster Detection}
Earthquake prediction relies heavily on earthquake precursor data. Anomalies associated with earthquake precursors can be classified into two main categories: tendency changes and high-frequency mutations. When a tendency does not follow its normal periodic evolution, it is called a changing tendency. Disturbance of high frequency refers to sudden changes in observations that occur with high frequency and large amplitude and often show irregular patterns. \citet{cai2019anomaly} develop a predictive model for normal data by employing LSTM units. Further advantages of LSTM networks include the ability to detect earthquake precursor data without elaborating preprocessing directly.

The detection of earthquakes in real time requires a high-density network to fully leverage inexpensive sensors. Over the past few years, low-cost acceleration sensors have become widely used for accurate earthquake detection. Accordingly, Petrol et al. \citep{perol2018convolutional} proposed CNNs for detecting earthquakes and locating them from two local stations in Oklahoma. Using deep CNNs, Phasenet \citep{zhu2019phasenet} is able to determine the arrival time of earthquake waves in archives.
In CrowdQuake \citep{huang2020crowdquake}, a convolutional RNN model is proposed as the core detection algorithm. Moreover, past acceleration data can be stored in databases and analysed post-hoc to identify earthquakes that may have been missed by real-time detection. In this model, abnormal sensors that might compromise earthquake events can be identified regularly.

\subsection{Energy}
It is inevitable that purification and refinement will affect various petroleum products. Regarding this, \citep{filonov2016multivariate} as an LSTM-based approach is employed to monitor and detect faults in a multivariate industrial time series that includes signals from sensors and control systems of gasoil plant heating loop (GHL). Likewise, according to \citet{wen2019time}, a CNN is used to detect time series anomalies using a transfer learning framework to solve data sparsity problems. The results were demonstrated on the GHL dataset \citep{filonov2016multivariate}, which contains data on cyber-attacks against utility systems. 

The use of phasor measurement units (PMU) by utilities for power system monitoring increases the potential for cyberattacks. In \citep{basumallik2019packet}, anomalies are detected in MTS data generated by PMU data packets corresponding to different events, such as line faults, trips, generation and load before each state estimation cycle. Consequently, it can help operators identify targeted cyber-attacks and make better decisions to ensure grid reliability.

The management of energy in buildings can improve energy efficiency, increase equipment life, and reduce energy consumption and operational costs. \citet{fan2018analytical} propose an autoencoder-based ensemble method for the analysis of energy time series in buildings and the detection of unexpected consumption patterns and excessive waste.

\subsection{Industrial Control Systems}
System calls can be generated through regularly scheduled tasks, which are a consequence of events from a given process, and sometimes, they are caused by interrupts that are triggered by events. It is difficult to construct profiles using system call information since some processes are time-driven, event-driven, or both. 

THREAT \citep{ezeme2020framework} provides a deeper insight into anomaly detection in system processes using their properties and system calls. Detecting anomalies at the kernel level provides new insights into the more complex machine-to-machine interactions. This is achieved by extracting useful features from system calls to detect a broad scope of anomalies.

An AE based on LSTM was implemented by \citet{hsieh2019unsupervised} to detect anomalies in multivariate streams occurring in production equipment components. In this technique, LSTM networks are used to encode and decode actual values and evaluate deviations between reconstructed and actual values.
Using CNN to handle MTS generated from semiconductor manufacturing processes is the basis for the model in \citep{kim2019fault}. Further, an MTS-CNN is proposed in \citep{hsu2021multiple} to detect anomalous wafers and provide useful information for root cause analysis in semiconductor production.

\subsection{Robotics}
In the modern manufacturing industry, as production lines become increasingly dependent on robots, failures of any robot can cause a plunge into a disastrous situation, while some faults are difficult to identify. In order to detect incipient failures in robots before they stop working completely, a real-time method is required to continuously track robots by collecting time series from robots. A sliding-window convolutional variational autoencoder (SWCVAE) is proposed in \citep{chen2020unsupervised} to detect anomalies in MTS both spatially and temporally in an unsupervised manner.

Also, many people with disabilities require physical assistance from caregivers, although robots can substitute some human caregiving. Robots can help with daily living activities, such as feeding and shaving. By detecting and stopping abnormal task execution in assistance, potential hazards can be prevented or reduced \citep{park2018multimodal}.

\subsection{Environmental Management}
In ocean engineering, structures and systems are designed in or near the ocean, such as offshore platforms, piers and harbours, ocean wave energy conversion, and underwater life-support systems. The ocean observing system (OOS) provides marine data by using sensors and equipment that work under severe conditions. In order to prevent big losses from total machine failure or even natural disasters, it is necessary to detect OOS anomalies early enough. The real-time OceanWNN model \citep{wang2019study}  leverages a novel WNN-based (Wavelet Neural Network) method for detecting anomalies in ocean fixed-point observing time series without any labelled training data. Wastewater treatment plants (WWTPs) play a crucial role in protecting the environment. A method based on LSTMs was used by \citep{mamandipoor2020monitoring} to monitor the process and detect collective faults in WWTPS, superior to earlier methods. Moreover, energy management systems must manage gas storage and transportation continuously in order to reduce expenses and safeguard the environment. In \citep{song2021gas}, an end-to-end CNN-based model is used to implement an internal-flow-noise leak detector in pipes. 

\end{document}